\documentclass[11pt,a4paper]{article}
\usepackage{times,latexsym}
\usepackage{url}
\usepackage[T1]{fontenc}

\usepackage{CJKutf8}
\usepackage{graphicx}
\usepackage{makecell}
\usepackage{subcaption}
\usepackage{booktabs}
\usepackage{tabularx}
\usepackage{todonotes}
\usepackage{arydshln}
\usepackage{multirow}
\usepackage{makecell}
\usepackage{float}
\usepackage{amsfonts}
\usepackage{amsmath}
\usepackage{linguex}
\usepackage{enumitem}

\alignSubExtrue

\usepackage{colortbl}

\newcommand{\bluenew}[1]{\textcolor{blue}{#1}}

\newcommand{\zb}[0]{ZhoBLiMP}

\usepackage[acceptedWithA]{tacl2021v1}

\usepackage{xspace,mfirstuc,tabulary}

\newif\iftaclinstructions
\taclinstructionsfalse 
\iftaclinstructions
\renewcommand{\confidential}{}
\renewcommand{\anonsubtext}{(No author info supplied here, for consistency with
TACL-submission anonymization requirements)}
\newcommand{\instr}
\fi

\iftaclpubformat 

\else

\fi

\title{A Systematic Assessment of Language Models \\
with Linguistic Minimal Pairs in Chinese}

\author{
 \textbf{Yikang Liu\textsuperscript{1}\thanks{$^{\ast}$Work done during an internship at Tongyi Lab. $^{\#}$Corresponding authors}},
 \textbf{Yeting Shen\textsuperscript{1}},
 \textbf{Hongao Zhu\textsuperscript{1,5}},
 \textbf{Lilong Xu\textsuperscript{4}},
 \textbf{Zhiheng Qian\textsuperscript{1}},
 \textbf{Siyuan Song\textsuperscript{1,3}},
\\
 \textbf{Kejia Zhang\textsuperscript{1}},
 \textbf{Jialong Tang\textsuperscript{2}},
 \textbf{Pei Zhang\textsuperscript{2}},
 \textbf{Baosong Yang\textsuperscript{2}},
 \textbf{Rui Wang\textsuperscript{1\#}},
 \textbf{Hai Hu\textsuperscript{1,6\#}}
\\
 \textsuperscript{1}Shanghai Jiao Tong University, China
\quad{}
 \textsuperscript{2}Tongyi Lab, Alibaba Group, China
\\
 \textsuperscript{3}The University of Texas at Austin, USA
\quad{}
 \textsuperscript{4}University of Glasgow, UK
\\
 \textsuperscript{5}University of California, San Diego, USA 
\quad{}
 \textsuperscript{6}City University of Hong Kong, China
\\
 \small{
   \textbf{Correspondence:} \href{mailto:email@domain}{yikangliu@sjtu.edu.cn}; 
   \href{mailto:email@domain}{wangrui12@sjtu.edu.cn};
   \href{mailto:email@domain}{hu.hai@outlook.com}
 }
}

\date{}

\begin{document}
\begin{CJK*}{UTF8}{gkai}

\maketitle
\begin{abstract}
We present ZhoBLiMP, the largest linguistic minimal pair benchmark for Chinese, with over 100 paradigms, ranging from topicalization to the \textit{Ba} construction. We then train from scratch a suite of Chinese language models (LMs) with different tokenizers, parameter sizes, and token volumes, to study the learning curves of LMs on Chinese. To mitigate the biases introduced by unequal lengths of the sentences in a minimal pair, we propose a new metric named sub-linear length normalized log-probabilities (SLLN-LP). Using SLLN-LP as the metric, our results show that \textsc{Anaphor}, \textsc{Quantifiers}, and \textsc{Ellipsis} in Chinese are difficult for LMs even up to 32B parameters, and that SLLN-LP successfully mitigates biases in ZhoBLiMP, JBLiMP and BLiMP. We conclude that future evaluations should be more carefully designed to consider the intricate relations between linking functions, LMs, and targeted minimal pairs.


\end{abstract}

\section{Introduction}
\label{sec:intro}

Acceptability judgment is an important empirical method to measure human linguistic competence~\citep{chomsky1965aspects,schutze2016empirical}, which has also been used to assess the linguistic knowledge of language models (LMs).
Much work in this line adopted the minimal pair paradigm (\textsc{mpp}) in evaluating LMs~\citep{linzen2016assessing,wilcox2018rnn,warstadt-et-al-2020-blimp,hu-levy-2023-prompting,warstadt-etal-2023-findings-babylm}.
A minimal pair is a pair of sentences with minimal contrast that affects whether the sentence is acceptable or not. A well-trained LM should assign higher probabilities to the acceptable sentences than the unacceptable ones (marked with *): 

\ex. \label{ex:intro-mp}
\a. The bureaucrat was bribed deliberately.
\b. *The bureaucrat bribes deliberately.\footnote{Taken from \citet[][pp.~239]{sprouse2013comparison-linguistic-inquiry}.}

Well-curated and large-scale \textsc{mpp} benchmarks are some of the most widely used benchmarks for assessing LMs' linguistic competence due to their ease of use~\citep{warstadt-etal-2023-findings-babylm,alkhamissi2025language-cognition-llms-out-grow}, and have also been widely used to study 
the mechanisms of language acquisition in LMs. 
For instance, using benchmarks such as the  English BLiMP~\citep[Benchmark of Linguistic Minimal Pair,][]{warstadt-et-al-2020-blimp}, LMs have been found to acquire syntax with just around 100M tokens~\citep{zhang-etal-2021-need,warstadt-etal-2023-findings-babylm}, show similar acquisition order regardless of initialization, architecture and training data~\citep{choshen-etal-2022-grammar}, and sometimes over-generalize with U-shaped learning patterns in which models only truly acquire a linguistic phenomenon after an initial dip in performance~\citep{evanson-etal-2023-language,haga-etal-2024-modeling}.

These findings were mostly English-centric, highlighted by the use of the BLiMP. While recent endeavors in other languages---JBLiMP for Japanese~\citep{someya2023jblimp}, BLiMP-NL for Dutch~\citep{blimp-nl}, among others---have facilitated research in more languages~(see Table~\ref{tab:intro:blimps}), there is still no systematic study of LMs' learning patterns in any non-English language. 

For one of the most widely spoken languages, Chinese, there are two existing \textsc{mpp} benchmarks: CLiMP~\citep{xiang-et-al-2021-climp} and SLING~\citep{song2022sling}.
However, both
fall short of including enough linguistic phenomena, with CLiMP only covering 16 and SLING 38.
Furthermore, CLiMP uses a lexicon translated from English, which has been noted for generating unnatural sentences~\citep{song2022sling}. 
Sentences in SLING are derived from the Penn Chinese Treebank~\cite{xue2005penn}, which limits its sentence structures to mainly the news domain and also makes it difficult to extend to new paradigms.

On the other hand, an implicit constraint for all BLiMP-style benchmarks is that the two sentences in a minimal pair should have equal lengths, for unbiased calculation of log-probabilities.
Yet this is difficult to observe, since in linguistic research for human acceptability judgment it is common for one sentence to have more/fewer words, as shown in \ref{ex:intro-mp}. 
Even if the two sentences had the same number of words, the sentences might be tokenized with different numbers of tokens due to subword tokenization. 
\citet{ueda2024token-length-bias} argued to filter out pairs of unequal length, but in our opinion, it would be better to design metrics that normalize length-related biases
to allow more flexibility in data curation.


\begin{table}[t]
\resizebox{\linewidth}{!}{
\begin{tabular}{lllr}
    \toprule
    Benchmark & Language & Size & N \\\midrule
    BLiMP~\cite{warstadt-et-al-2020-blimp} & English & 67k & 67 \\
    SyntaxGym~\cite{hu-etal-2020-systematic-assessment-syn-gen} & English & NA & 39 \\
    CLiMP~\cite{xiang-et-al-2021-climp} & Chinese & 16k & 16 \\
    SLING~\cite{song2022sling} & Chinese & 38k & 38 \\
    JBLiMP~\cite{someya2023jblimp} & Japanese & 331 & 39 \\
    \multirow{2}{*}{LINDSEA~\cite{leong2023bhasa}} & Indonesian & 380 & 38  \\ 
     & Tamil & 200 & 20  \\
    RuBLiMP~\cite{rublimp} & Russian & 45k & 45  \\
    BLiMP-NL~\citep{blimp-nl} & Dutch & 8.4k & 84 \\\midrule
    ZhoBLiMP (Ours) & Chinese & 35k & 118 \\\bottomrule
    \end{tabular}}
    \caption{Comparison of \textsc{mpp} benchmarks for different languages. \textit{Size} refers to the number of minimal pairs in total; \textit{N} refers to the number of linguistic paradigms.}
\label{tab:intro:blimps}
\end{table}

\begin{table*}[ht]
    \centering
    \resizebox{0.99\linewidth}{!}{
    \begin{tabular}{llll} \toprule
        Phenomenon & N & Acceptable example & Unacceptable example \\\midrule
        \multirow{2}{*}{\textsc{Anaphor}} & \multirow{2}{*}{6} & \footnotesize{\textit{她的弟弟讨厌\bluenew{他自己}。}} & \footnotesize{\textit{她的弟弟讨厌\bluenew{她自己}。}} \\
         & & \footnotesize{\textit{Her little brother hates \bluenew{himself}.}} & \footnotesize{\textit{Her little brother hates \bluenew{herself}.}} \\
        \multirow{2}{*}{\textsc{Argument struc.}} & \multirow{2}{*}{7}  & \footnotesize{\textit{我\bluenew{预习}了教材。}} & \footnotesize{\textit{我\bluenew{出现}了教材。}} \\
         & & \footnotesize{\textit{I \bluenew{previewed} the textbook.}} & \footnotesize{\textit{I \bluenew{appeared} the textbook.}}\\
        \multirow{2}{*}{\textsc{BA}} & \multirow{2}{*}{13}  & \footnotesize{\textit{她把那条鱼放在池塘里。}} & \footnotesize{\textit{把那条鱼她放在池塘里。}} \\
         & & \footnotesize{\textit{She BA that fish put in the pond.}} & \footnotesize{\textit{BA that fish she put in the pond.}}\\
        \multirow{2}{*}{\textsc{Classifier}} & \multirow{2}{*}{3}  & \footnotesize{\textit{那边站着八\bluenew{位}舞者。}} & \footnotesize{\textit{那边站着八\bluenew{条}舞者。}} \\
         & & \footnotesize{\textit{Eight WEI dancers are standing there.}} & \footnotesize{\textit{Eight TIAO dancers are standing there.}}\\
        \multirow{2}{*}{\textsc{Cntl \& Raising}} & \multirow{2}{*}{4}  & \footnotesize{\textit{那杯红酒\bluenew{会}变质。}} & \footnotesize{\textit{\bluenew{会}那杯红酒变质。}} \\
         & & \footnotesize{\textit{That glass of wine \bluenew{will} go bad.}} & \footnotesize{\textit{\bluenew{Will} that glass of wine go bad.}}\\
        \multirow{2}{*}{\textsc{Ellipsis}} & \multirow{2}{*}{3}  & \footnotesize{\textit{你们\bluenew{拉}了小提琴，我们也\bluenew{拉}了。}} & \footnotesize{\textit{你们\bluenew{笑}了一天，我们也\bluenew{笑}了。}} \\
         & & \footnotesize{\textit{You \bluenew{played} the violin, we \bluenew{played} too.}} & \footnotesize{\textit{You \bluenew{laughed} all day, we \bluenew{laughed} too.}}\\
        \multirow{2}{*}{\textsc{FCI licensing}}  & \multirow{2}{*}{5}  & \footnotesize{\textit{任何人\bluenew{都}可以去。}} & \footnotesize{\textit{任何人可以去。}} \\
         & & \footnotesize{\textit{Anyone can go.}} & \footnotesize{\textit{Anyone go.}}\\
        \multirow{2}{*}{\textsc{Nominal exp.}}  & \multirow{2}{*}{11}  & \footnotesize{\textit{他\bluenew{是}司机。}} & \footnotesize{\textit{他司机。}} \\
         & & \footnotesize{\textit{He is a driver.}} & \footnotesize{\textit{He driver.}}\\
        \multirow{2}{*}{\textsc{NPI licensing}} & \multirow{2}{*}{9}  & \footnotesize{\textit{没有任何人来了。}} & \footnotesize{\textit{任何人没有来了。}} \\
         & & \footnotesize{\textit{Nobody came.}} & \footnotesize{\textit{Anyone didn’t come.}}\\
        \multirow{2}{*}{\textsc{Passive}} & \multirow{2}{*}{12}  & \footnotesize{\textit{他被小明打断了\bluenew{鼻子}。}} & \footnotesize{\textit{他被小明打断了\bluenew{教材}。}} \\
         & & \footnotesize{\textit{His \bluenew{nose}  was hit-broken by Xiao Ming.}} & \footnotesize{\textit{His \bluenew{textbook} was breathed hit-broken by Xiao Ming.}}\\
        \multirow{2}{*}{\textsc{Quantifiers}} & \multirow{2}{*}{2}  & \footnotesize{\textit{没有人吃了\bluenew{超过}九块糖果。}} & \footnotesize{\textit{没有人吃了\bluenew{至少}九块糖果。}} \\
         & & \footnotesize{\textit{No one ate \bluenew{more than} nine candies.}} & \footnotesize{\textit{No one ate \bluenew{at least} nine candies.}}\\
        \multirow{2}{*}{\textsc{Question}} & \multirow{2}{*}{21}  & \footnotesize{\textit{你\bluenew{到底}喝不喝啤酒？}} & \footnotesize{\textit{你\bluenew{难道}喝不喝啤酒？}} \\
         & & \footnotesize{\textit{You DAODI will drink the beer or not?}} & \footnotesize{\textit{You NANDAO will drink the beer or not?}}\\
        \multirow{2}{*}{\textsc{Relativization}} & \multirow{2}{*}{4}  & \footnotesize{\textit{我知道赵大爷不想要\bluenew{你}笑的原因。}} & \footnotesize{\textit{我知道赵大爷不想要\bluenew{谁}笑的原因？}} \\
         & & \footnotesize{\textit{I know why Zhao doesn't want you to laugh.}} & \footnotesize{\textit{I know why Zhao doesn't want who to laugh?}}\\
        \multirow{2}{*}{\textsc{Topicalization}} & \multirow{2}{*}{4}  & \footnotesize{\textit{我们没盖\bluenew{什么被子}。}} & \footnotesize{\textit{我们\bluenew{什么被子}没盖。}} \\
         & & \footnotesize{\textit{We didn't have any quilt.}} & \footnotesize{\textit{We any quilt didn't have.}}\\
        \multirow{2}{*}{\textsc{Verb Phrase}} & \multirow{2}{*}{14}  & \footnotesize{\textit{她没有吃\bluenew{过}蛋糕。}} & \footnotesize{\textit{她没有吃\bluenew{了}蛋糕。}} \\
         & & \footnotesize{\textit{She hasn’t eaten a cake.}} & \footnotesize{\textit{She hasn’t ate a cake.}}\\ \bottomrule
    \end{tabular}}
    \caption{Overview of the 15 phenomena in ZhoBLiMP, with the number of paradigms (N), and one randomly sampled minimal pair. Each paradigm contains 300 minimal pairs. English translation for illustrative purposes, using a mixture of word-by-word gloss and translation to show the contrast. }
    \label{tab:intro:examples}
\end{table*}

The research gaps above are twofold: (1) the relative inadequacy in resources in Chinese, and (2) the challenges posed by length-related bias in evaluation.
To fill these gaps, we present ZhoBLiMP, train the Zh-Pythia LM suite, and propose a new linking function to debias model evaluation.

\zb{} is the largest \textsc{mpp} benchmark for Chinese to date, with 118 paradigms\footnote{``Paradigm'' refers to the patterns of \textit{one} minimal pair. 118 paradigms means 118 such minimal pair patterns.} covering 15 linguistic phenomena, totaling 35k minimal pairs (see Table~\ref{tab:intro:examples} for examples). 
Compared with CLiMP and SLING, we consult more Chinese linguistic literature to obtain a more comprehensive coverage, particularly phenomena frequently discussed in the field of Chinese syntax, such as classifiers, the \textit{ba} and \textit{bei} constructions. Besides, we loosen the ``equal length'' constraint, without controlling the length at all levels, including word, character, or token, which allows us to include phenomena such as \textit{ellipsis} (\S\ref{sec:zhoblimp}).

We then train a suite of transformer-based LMs of different sizes (14M-1.4B parameters) from scratch on Chinese text (10M-3B tokens), with three types of tokenization methods and a careful checkpointing strategy, to study the learning patterns and trajectories of LMs, as well as the effect of tokenization on model performance (\S\ref{sec:zh-pythia}).

With ZhoBLiMP and the LM suite, we quantify the biases caused by unequal lengths and propose a new linking function---sub-linear length normalized log-probabilities (SLLN-LP)---to mitigate the bias. 
Through extensive validation on Chinese and preliminary experimentation on English, Dutch and Japanese, we find that SLLN-LP can successfully mitigate the ``unequal length'' problem and improve general accuracy in ZhoBLiMP, BLiMP, and JBLiMP, potentially applicable to other languages as well (\S\ref{sec:debias}).

Using SLLN-LP to assess the trained LM suite and the Qwen2.5 series on ZhoBLiMP, we find that an LM of 160M parameters trained on 3B tokens can achieve $\sim$87\% accuracy on par with much larger multilingual LLMs.
Notably, there are three phenomena that are challenging for LMs even up to 14B parameters: \textsc{Anaphor}, \textsc{Ellipsis}, and \textsc{Quantifiers} (\S\ref{sec:assess}). 
We discuss our findings in the context of previous BLiMP benchmarks and provide suggestions for future researchers in \S\ref{sec:discussion}.


Minimal pairs in ZhoBLiMP, our codebase for data generation and the Zh-Pythia suite are available 
at \url{https://github.com/sjtu-compling/ZhoBLiMP}. 


\section{Creation of ZhoBLiMP}
\label{sec:zhoblimp}

Bearing in mind the gaps in current benchmarks, we adopt the strategy of generation from templates and a vocabulary, to build a controlled and extendable Chinese benchmark of minimal pairs. 
To do so, we first build a GUI that can generate minimal pairs given grammar templates of a minimal pair and a vocabulary. 
Then, Chinese linguists manually write the grammar templates for minimal pairs extracted from multiple sources
and create a vocabulary with the necessary features. 
Finally, we hire Chinese native speakers to validate samples in ZhoBLiMP.

\subsection{Minimal pair generation platform}
\label{sec:zhoblimp:platform}

\paragraph{Features of the platform.} 

\begin{figure}
    \centering
    \includegraphics[width=\linewidth]{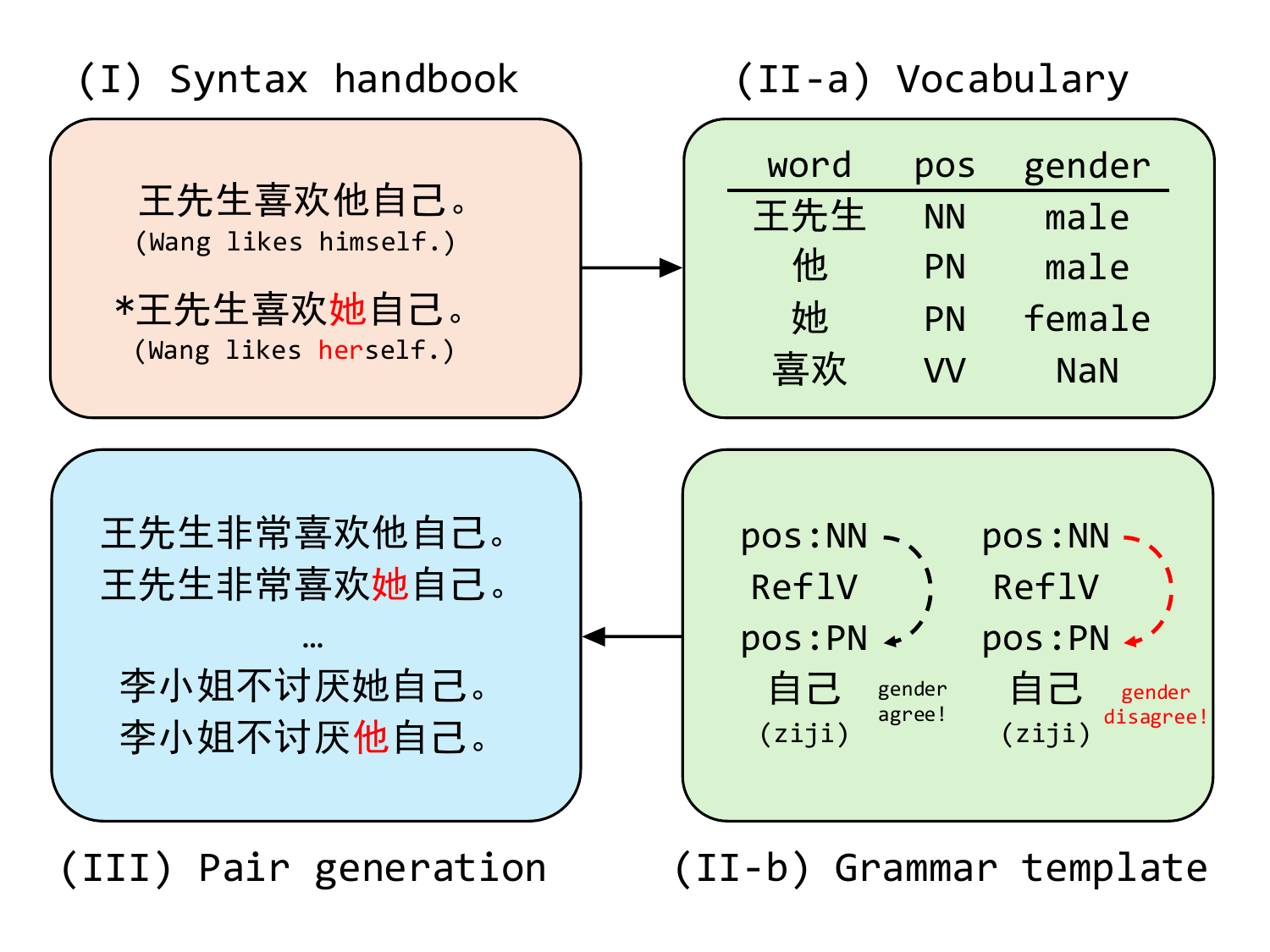}
    \caption{Data generation procedure illustration. Without writing any codes, linguists can easily generate sentence pairs by crafting grammar templates and vocabulary in the green blocks.}
    \label{fig:zhoblimp:data-gen}
\end{figure}

The platform provides users with a web interface to craft grammar templates with four types of rules listed below to generate minimal pairs (see Figure~\ref{fig:zhoblimp:data-gen}). 

\begin{itemize}[leftmargin=*]
\setlength{\itemsep}{0em}
\setlength{\parskip}{0em}
    \item \texttt{\textbf{Lexical}}: A set of key-value pairs assigning values to certain lexical properties. Lexical items will be searched accordingly in the vocabulary. E.g., \texttt{pos:NN} will randomly sample lexical items the part-of-speech of which is \texttt{NN}.
    \item \texttt{\textbf{Direct}}: A list of string expressions that can be directly used in the composition of sentences rather than searching the vocabulary. E.g., ``\textit{自己}'' (\textit{ziji}) will directly occupy the position. 
    \item \texttt{\textbf{(mis)Matched}}: Similar to \texttt{Lexical}, but assigns (dis)agreement in one lexical property between two different positions. E.g., \texttt{pos:PN mPos:0 mPro:gender}\footnote{Stands for: POS is pronoun; matched position is 0; matched property is gender. } samples an item of PN that agrees in gender with the first item in the template. 
    \item \texttt{\textbf{Phrase}}: A pre-defined phrase that supports recursion of various depths. \texttt{ReflV} yields a phrase such as ``\textit{不喜欢}'' or ``\textit{非常喜欢}'' (\textit{don't like} or \textit{very like}). 
\end{itemize}

A grammar template consists of a good grammar and a bad one, both containing several rules.  
With this interface, users can easily generate minimal pairs at scale without any coding.

\paragraph{Vocabulary and grammar template.}
We design lexical properties that can be utilized in the lexicon filtering, and transform linguistic constraints into rule expressions according to lexical properties. 
As illustrated in the minimal pair:

\ex. \label{ex:data-gen}
\ag. \textit{王先生}$_i$ \textit{非常} \textit{喜欢} \textit{他自己}$_i$。\\
Mr.~Wang$_i$ very likes himself$_i$. \\
\bg. *\textit{王先生}$_i$ \textit{非常} \textit{喜欢} \textit{她自己}$_i$。 \\
Mr.~Wang$_i$ very likes herself$_i$. \\

The ungrammaticality in \ref{ex:data-gen} is caused by gender disagreement between \textit{王先生} (Mr. Wang) and \textit{她自己} (herself), which are co-indexed. 
We can use \texttt{(mis)Matched} to create the gender (dis)agreement between the reflexive (\texttt{pos:PN} + \textit{ziji}, \textit{她}/\textit{他}+\textit{自己}) and its antecedent (\texttt{pos:NN}, \textit{王先生}), where the pronoun is a \texttt{Lexical} rule and the \textit{ziji} (\textit{self}) is a \texttt{Direct} rule. 
To complete the sentence structure, we use the phrase \texttt{ReflV}, which is pre-defined as a predicate that can take the reflexive pronoun as an object.

\subsection{Data generation for ZhoBLiMP}
\label{sec:zhoblimp:data-gen}

\paragraph{Sources of the minimal pairs.}
Templates of minimal pairs, which we call paradigms, come from three sources: (1) minimal pairs in a syntax textbook on Chinese---\textit{The Syntax of Chinese}~\cite{SyntaxHLL}\footnote{We use large-scale native speaker ratings of these minimal pairs reported in \citet{chen2020assessing} to select those pairs where judgments of linguists converge with those of native speakers.  }, (2) BLiMP~\cite{warstadt-et-al-2020-blimp}, and (3) journal articles on Chinese syntax and linguistics. 
We choose (1) because as a general-purpose textbook, it provides a comprehensive and systematic description of almost all syntactic phenomena in Chinese, with many example minimal pairs, which ensures ZhoBLiMP's coverage. 
For phenomena not covered in (1), we manually selected linguistics articles (3) that discuss these phenomena, and extract minimal pairs from them. 
We also select paradigms in the English BLiMP that have counterparts in Chinese for future comparison of the two languages. 

\paragraph{Generation from templates and a vocab.}
A group of eight Chinese linguists first extract the minimal pairs from the three sources above. 
Then grammar rules for each sentence in the minimal pair are created, and the necessary lexical items with their specific features are added to the vocabulary, which are then used to generate 300 unique minimal pairs for each paradigm. 
We consult linguistic resources on specific phenomena to create a wide coverage vocabulary that can cover the 100+ paradigms we generate. The creation of templates and vocabulary took about two months.  

In the end, 118 paradigms are created, which are grouped into 15 linguistic phenomena, with examples listed in Table~\ref{tab:intro:examples}.

\subsection{Data Validation}
\label{sec:zhoblimp:valid}

We conduct human validation to verify the quality of the generated minimal pairs.
We randomly sample 5 minimal pairs from each paradigm, and create a questionnaire asking native speakers which sentence from the minimal pair sounds more natural. 
All sampled minimal pairs were split into 10 lists, each containing roughly 65 pairs, plus two catch trials that a participant must answer correctly for her data to be valid.
Fifty native speakers of Chinese are hired, each completing one list, leading to 5 responses per minimal pair. 
Participants are rewarded 10 Chinese RMB
for the task.
This validation shows that the participants have an overall agreement of 93.9\% with our gold labels, with the agreement for each phenomenon listed in Table~\ref{tab:assess:overall}.
This shows our benchmark aligns with the intuition of Chinese native speakers. 

\section{Training Zh-Pythia LM suite}
\label{sec:zh-pythia}

We train a series of LMs from scratch in a corpus mainly consisting of Chinese texts to investigate the acquisition of Chinese grammar with different configurations: number of model parameters, size of training corpora, and tokenizers. We name this suite of LMs as Zh-Pythia after Pythia~\citep{biderman2023pythia}, borrowing their model architecture and configurations.\footnote{Models are released at \url{https://huggingface.co/collections/SJTU-CL/zh-pythia}.}

For the training corpus, we collect 7000+ books that are primarily in the fields of humanities, including fiction and non-fictional topics on history, psychology, etc. The full corpus takes up 12GB in the txt format. 
We train LMs from scratch with different amounts of Chinese texts (see Table~\ref{tab:zh-pythia:config}), resulting in 20 different combinations. 

\begin{table}[ht]
    \centering
    \begin{tabular}{l|c}
    \toprule
        \# params & 14M,70M,160M,410M,1.4B \\
        \# tokens & 10M,100M,1B,3B \\
    \bottomrule
    \end{tabular}
    \caption{Number of model parameters and training tokens of Zh-Pythia language models.}
    \label{tab:zh-pythia:config}
\end{table}

We use off-the-shelf Chinese-Llama~(\textit{cllama}, \citealp{chinese-llama-alpaca}) as our main tokenizer. \citet{chinese-llama-alpaca} create a Chinese vocabulary with 20k items and then merge it with the Llama tokenizer~\citep{llama-2023}. 
We train models with all configurations listed in Table~\ref{tab:zh-pythia:config} with \textit{cllama}. 
In addition to that, we train two more tokenizers from scratch on our training corpus. 
One is a character-level tokenizer. The other is word-level, using characters as the base vocabulary and merging them into multi-character words with the BPE algorithm~\citep{gage1994new,sennrich-etal-2016-neural} (see Table~\ref{tab:zh-pythia:tokenizers}). We additionally train two more LMs of 160M parameters on 3B tokens using the two tokenizers, respectively.

\begin{table}[ht]
\small
    \centering
    \begin{tabular}{c|ccc}
    \toprule
        tokenizer & \# all & \# single-char & \# multi-char \\\midrule
        \textit{cllama} & 49,953 & 10,876 & 6,963 \\
        \textit{word} & 20,276 & 11,116 & 7,471 \\
        \textit{char} & 12,142 & 11,116 & 0\\
    \bottomrule
    \end{tabular}
    \caption{Tokenizers used for model training: \textit{all} denotes the total size of the vocabulary; \textit{single-char} and \textit{multi-char} only include tokens that contain Chinese characters.}
    \label{tab:zh-pythia:tokenizers}
\end{table}

We set the global batch size at 256 $\times$ 64 tokens for all models and keep the hyperparameters the same. All LMs are trained on three random seeds.
Training is done with two A100-80G GPUs.

\section{Debiasing length normalization in minimal pairs}
\label{sec:debias}

In this section, we first briefly review how existing linking functions are developed, and show that raw or mean-log probabilities of LMs will bias the evaluation when two sentences in a pair are different in length. We then introduce and validate a new function that addresses the unequal length problem by performing sub-linear length normalization. 

\subsection{Baseline linking functions}
\label{sec:debias:funcs}

Token probabilities assigned by LMs provide an unsupervised signal for acceptability judgment: higher probabilities indicate greater acceptability. While LMs themselves are often evaluated on acceptability judgment tasks, their probabilities can also be used to predict human acceptability ratings. Linking functions bridge this gap between LM probabilities and human acceptability ratings, requiring validity on human Likert-scale ratings.

The most basic linking function is raw log-probability (LP). For a tokenized sequence $x$ with $|x|$ tokens:
\begin{align}
    \text{LP}(x) = \sum_{i=1}^{|x|} \log P(x_i|x_{<i}) \notag
\end{align}
\citet{lau-et-al-2017-grammar} demonstrate that normalizing for length (\textit{mean LP}, MLP) and lexical unigram frequency (\textit{syntactic log-odds ratio}, SLOR) correlates better with human ratings:
\begin{align*}
    \text{MLP}(x) &= \frac{\text{LP}(x)}{|x|}. \\\notag
    \text{SLOR}(x) &= \frac{\text{LP}(x) - \text{U}(x)}{|x|}. \notag
\end{align*}
where $\text{U}(x) = \sum_{i}^{|x|} \log P_{u}(x_i)$ and $P_{u}(x_i)$ represent the token frequency in the training corpus. More recently, \citet{lau2020-penlp} proposed Pen LP as an additional measurement under length normalization, based on \citet{wu2016google-NMT}:
\begin{align*}
    \text{PenLP}(x) &= \frac{\text{LP}(x)}{((|x|+5)/(5+1))^\alpha}.
\end{align*}
where $\alpha$ usually takes the value of 0.8.
Additionally, \citet{tjuatja-etal-2025-goes-morcela} find that frequency normalization requirements vary across LMs and propose MORCELA with adjustable normalization:
\begin{align*}
    \text{MORCELA}(x) = \frac{\text{LP}(x) - \beta \cdot \text{U}(x) + \gamma}{|x|}. \notag
\end{align*}
where tunable parameters $\beta$ and $\gamma$ further strengthen the correlation. While length normalization in these functions aims to improve prediction for independent sentences, their application may differ in \textsc{mpp} contexts.

\subsection{Goal of debiasing length normalization}
\label{sec:debias:goal}

Our goal here is to find a linking function $f$ that can properly perform length normalization without over-penalizing longer or shorter sentences. 
We categorize a benchmark $\mathcal{D}$ consisting of minimal pairs $(g, u)$, where $g$ and $u$ refer to grammatical and ungrammatical sentences, into three parts:
\begin{align*}
    \mathcal{D}_{<} = {(g,u) \in \mathcal{D} : |g| < |u|} \\\notag
    \mathcal{D}_{=} = {(g,u) \in \mathcal{D} : |g| = |u|} \\\notag
    \mathcal{D}_{>} = {(g,u) \in \mathcal{D} : |g| > |u|} \notag
\end{align*}
where $\mathcal{D}_{<}$ consists of pairs, the grammatical sentence of which is shorter than the ungrammatical counterpart, while the grammatical sentences are longer than the ungrammatical ones in $\mathcal{D}_{>}$, and $\mathcal{D}_{=}$ is more standard minimal pairs of the same length.
We then estimate the linguistic knowledge of LMs by the accuracy of correctly selecting the good sentence from a minimal pair using the linking function $f$ given a dataset $\mathcal{D}$:
\begin{equation}
    \text{acc}(\mathcal{D}; f) = \frac{1}{|\mathcal{D}|}\sum_{(g,u)\in \mathcal{D}} \mathbb{I}(f(g)>f(u)). \notag
\end{equation}
An ideal $f$ will generate similar model predictions no matter how the length varies between two sentences in a pair. We take the accuracy on split $\mathcal{D}_{=}$ as reference, as these pairs are not affected by length normalization. If the linking function is robust to the length difference in a pair, then the performance on $\mathcal{D}_{<}$ and $\mathcal{D}_{>}$ should be close to the reference accuracy.
Thus, we formalize the goal of debiasing length normalization as $\Delta_{acc}$, and we want to find the $f$ that minimizes the value:
\begin{align}
    \Delta_{acc} = \frac{1}{2} \sum_{\mathcal{D} \in \{\mathcal{D}_{<},\mathcal{D}_{>}\}} |\text{acc}(\mathcal{D}; f) - \text{acc}(\mathcal{D}_{=}; f)|. \notag
\end{align}

\subsection{SLLN-LP: Sublinear length normalized log-probabilities}
\label{sec:debias:slln}

We observe that LP favors shorter sentences without normalization, while MLP sometimes over-normalizes longer sentences. For instance, Zh-Pythia-160m achieves only 60.77\% of accuracy on $\mathcal{D}_{>}$ but 95.62\% on $\mathcal{D}_{<}$ when using LP. In contrast, the accuracy on $\mathcal{D}_{>}$ improves to 96.78\% but the accuracy on $\mathcal{D}_{<}$ drops to 66.19\% (see Table~\ref{tab:debias:fc}), while Pen LP has a more balanced performance. We observe that the degrees of length normalization might be the solution. 
Thus, we propose to use a sublinear function, in between LP and MLP, to normalize length:
\begin{equation}
    \text{SLLN-LP}(x;\alpha) = \frac{\text{LP}(x)}{|x|^{\alpha}}, \text{where}~\alpha \in (0, 1). \notag
\end{equation}
In SLLN-LP, the parameter $\alpha$ determines the degree of length normalization. Higher values of $\alpha$ yield more aggressive normalization. The boundary cases of $\alpha = 0$ and $\alpha = 1$ reduce SLLN-LP to LP and MLP, respectively. With this parameter, we aim to find how much we should perform length normalization to mitigate the noise brought by pairs of unequal lengths.

We exclude frequency normalization in SLLN-LP for two reasons: (1) Frequency should not significantly impact forced-choice since minimal pairs mostly contain identical words. (2) We aim to isolate length normalization from frequency normalization. Without frequency normalization, LM log-probabilities in the numerators remain negative. Frequency normalization can alter these signs, which changes the monotonicity of length normalization.\footnote{Signs of over 90\% of sentences are altered using SLOR.}

\begin{table*}[ht]
\resizebox{\textwidth}{!}{
    \begin{tabular}{l|ccc|ccc|ccc|rr} 
    \toprule
         & \multicolumn{3}{c|}{\footnotesize Zh-Pythia-160m-cllama} & \multicolumn{3}{c|}{\footnotesize Zh-Pythia-160m-word} & \multicolumn{3}{c|}{\footnotesize Zh-Pythia-160m-char} & \\\midrule
        Length & $\mathcal{D}_{=}$ & $\mathcal{D}_{>}$ & $\mathcal{D}_{<}$ & $\mathcal{D}_{=}$ & $\mathcal{D}_{>}$ & $\mathcal{D}_{<}$ & $\mathcal{D}_{=}$ & $\mathcal{D}_{>}$ & $\mathcal{D}_{<}$ & \multirow{2}{*}{\textit{acc}$\uparrow$} & \multirow{2}{*}{$\Delta_{acc}$$\downarrow$} \\
        \# pairs & 20,584 & 5,456 & 9,360 & 22,206 & 5,802 & 7,392 & 23,347 & 6,198 & 5,855 & \\\midrule
        LP & 88.20 & {60.77} & 95.62 & 84.13 & {52.85} & 98.73 & \textbf{86.19} & {56.55} & 94.28 & 83.44 & 19.74 \\
        mean LP & 88.20 & 96.78 & {66.19} & 84.13 & 92.53 & {54.64} & \textbf{86.19} & 95.96 & {66.91} & 82.59 & 16.26 \\
        pen LP & 88.20 & 83.47 & 87.09 & 84.13 & 81.63 & 87.63 & \textbf{86.19} & 82.97 & 80.98 & 85.46 & \textbf{3.38} \\
        SLOR\textsuperscript{*} & 90.01 & 88.09 & 79.60 & \textbf{86.44} & 87.65 & 74.70 & 84.08 & 92.74 & {70.36} & 84.83 & 7.94 \\
        MORCELA\textsuperscript{*} & \textbf{90.12} & 82.32 & 88.48 & 86.12 & 83.55 & 82.90 & 86.15 & 91.03 & {74.88} & \textbf{86.22} & 5.23 \\\midrule
        SLLN-LP \\
        ~~~$\alpha=0.1$ & 88.20 & {66.07} & 93.66 & 84.13 & {60.09} & 97.79 & \textbf{86.19} & {61.85} & 91.78 & 84.04 & 15.87 \\
        ~~~$\alpha=0.3$ & 88.20 & {74.58} & 91.06 & 84.13 & {71.51} & 94.00 & \textbf{86.19} & {71.45} & 87.77 & 84.95 & 9.21 \\
        ~~~$\alpha=0.5$ & 88.20 & 81.60 & 87.55 & 84.13 & 81.18 & 87.66 & \textbf{86.19} & 79.85 & 82.93 & 85.31 & \underline{3.89} \\
        ~~~$\alpha=0.7$ & 88.20 & 89.11 & 80.90 & 84.13 & 87.79 & 77.60 & \textbf{86.19} & 87.72 & 76.46 & 84.87 & 4.94 \\
        ~~~$\alpha=0.9$ & 88.20 & 95.06 & {71.66} & 84.13 & 91.24 & {62.82} & \textbf{86.19} & 94.13 & {69.96} & 83.54 & 12.67 \\\bottomrule
    \end{tabular}}
    \caption{Accuracy (\textit{acc}) and length-related bias ($\Delta_{acc}$) on \zb{} using different acceptability linking functions. An appropriate function should have a higher \textit{accuracy} and a lower $\Delta_{acc}$ across tokenizers. More optimal parameters of MORCELA are searched independently of each LM (\textit{cllama}: $\beta=0.6$, $\gamma=18$; \textit{word}: $\beta=0.7$, $\gamma=12$; \textit{char}: $\beta=0.1$, $\gamma=12$). }
    \label{tab:debias:fc}
\end{table*}

\subsection{Validating SLLN-LP across tokenizers}
\label{sec:debias:validate}

In this and the next subsections, we comprehensively analyze the impact of length biases in Chinese, English, Japanese, and Dutch in more controlled settings. 

Different tokenization methods can result in varying sentence lengths. We first validate SLLN-LP's performance across different tokenizers.
We conduct experiments using three models: (1) Zh-Pythia-160m-cllama, (2) Zh-Pythia-160m-word, and (3) Zh-Pythia-160m-char. These models are trained with identical token volumes, with the tokenizer serving as the sole variable.
Using these differently tokenized models, we evaluate model performance
acceptability judgments 
through various linking functions (LP, mean LP, pen LP, SLOR, MORCELA, and SLLN-LP) on \zb{}.

\paragraph{Improper length normalization leads to bias in \textsc{mpp} evaluation.}

As shown in Table~\ref{tab:debias:fc}, approximately 30-40\% of pairs in ZhoBLiMP have unequal lengths, a proportion that remains consistent across different tokenizers. These unequal-length pairs are particularly sensitive to the choice of linking functions. As highlighted by the underlined cells in Table~\ref{tab:debias:fc}, improper normalization in linking functions can either underestimate or overestimate model performance, with accuracy varying significantly among $\mathcal{D}_{=}$, $\mathcal{D}_{+}$, and $\mathcal{D}_{-}$.

LP tends to assign higher log-probabilities to shorter sentences regardless of their grammaticality, resulting in poor performance on $\mathcal{D}_{>}$ but inflated performance on $\mathcal{D}_{<}$. Conversely, mean LP over-normalizes for length in \textsc{mpp}, favoring longer sentences due to their larger denominators, which leads to higher acceptability ratings. These opposing biases explain the substantial $\Delta_{acc}$ observed in both LP and mean LP methods.

\paragraph{Frequency normalized functions can mitigate the bias, but not always.}
The two starred functions in Table~\ref{tab:debias:fc} incorporate token frequency normalization. 
The results show that unigram frequency normalization substantially reduces $\Delta_{acc}$ to below 10 (specifically, 7.94 and 5.23) while improving overall average accuracy. 
For LMs trained with word-level tokenizers, this normalization also enhances accuracy on $\mathcal{D}_{=}$, where sentence pairs have equal lengths. However, LMs trained with character-level tokenizers still exhibit high length-difference bias and show accuracy drops on $\mathcal{D}_{=}$, producing results comparable to mean LP.
These findings suggest that frequency normalization effects are model-dependent, as evidenced by the variation in optimal frequency control parameters ($\beta$) of MORCELA across different models.

\begin{figure*}[t]
    \centering
    \includegraphics[width=\linewidth]{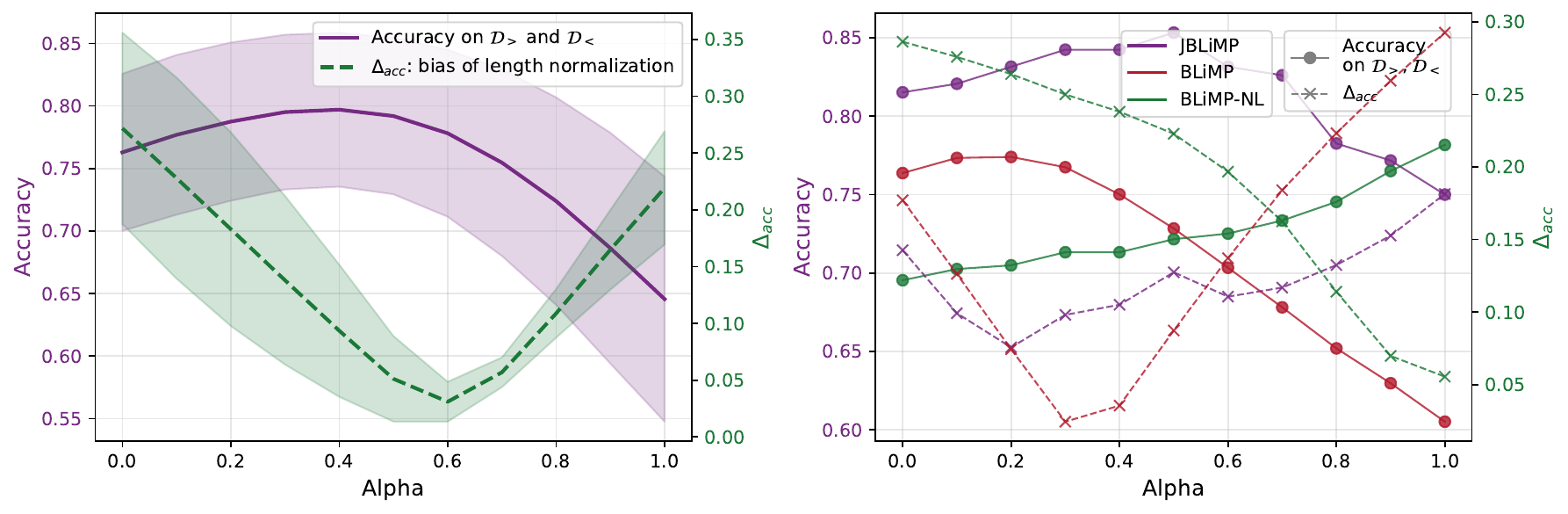}
    \caption{Effectiveness of debiasing length normalization using different $\alpha$. We report two metrics: (1) average \textit{accuracy} on $\mathcal{D}_{+}$ and $\mathcal{D}_{-}$ (when minimal pairs have unequal lengths), and (2) $\Delta_{acc}$.
    \textit{left} is the average results of 20 Zh-Pythia LMs with the shaded area denoting the standard deviation; \textit{right} is the results of \textsc{mpp} benchmarks in other languages, including JBLiMP, BLiMP, and BLiMP-NL.}
    \label{fig:debias:scale-language}
\end{figure*}

\paragraph{SLLN-LP normalizes length more effectively on \zb{}.} 

Our proposed SLLN-LP with $\alpha=0.5$ and pen LP demonstrate better length normalization, achieving the $\Delta_{acc}$ of 3.89 and 3.38, respectively. 
The accuracy obtained using these two functions is just slightly lower than that of MORCELA, which has more tunable parameters. 

As $\alpha$ in SLLN-LP increases from 0.1 to 0.9, we observe that $\Delta_{acc}$ initially decreases before increasing again. Correspondingly, \textit{accuracy} improves as the bias is gradually mitigated through more optimal values of the controlling exponent $\alpha$. 
These results suggest that the length-related bias in \zb{} can be mitigated by adjusting the degree of length normalization. Pen LP and SLLN-LP are both sub-linear and achieve equivalent performance, but we will use SLLN-LP below for its simpler conceptualization.

\subsection{Finding the optimal $\alpha$ of SLLN-LP across scales and languages}
\label{sec:debias:scale-language}

\begin{table*}[ht]
    \centering
    \resizebox{0.9\textwidth}{!}{
    \begin{tabular}{l|ccccc|cc|rr}
    \toprule
     & \multicolumn{5}{c|}{Zh-Pythia} & \multicolumn{2}{c|}{Qwen2.5} & Gap & Human \\\midrule
        Phenomenon & 14M & 70M & 160M & 410M & 1.4B & 7B & 14B &  &  \\\midrule
        \textsc{BA} & 90.37 & 91.19 & 95.91 & 96.39 & \textbf{96.65} & 92.28 & 91.03 & -0.50 & 96.15 \\
        \textsc{Anaphor} & 46.15 & 40.72 & 44.28 & 56.67 & 63.20 & 59.22 & \textbf{64.72} & \underline{20.87} & 85.60 \\
        \textsc{Arg. Structure} & 76.71 & 85.27 & 87.08 & 88.44 & \textbf{88.95} & 88.38 & 87.00 & 6.97 & 95.92 \\
        \textsc{Classifier} & 58.37 & 70.30 & 84.33 & 90.26 & \textbf{94.11} & 91.67 & 89.33 & -0.06 & 94.05 \\
        \textsc{Control Raising} & 86.67 & 92.67 & 97.36 & 97.86 & \textbf{98.28} & 91.25 & 94.92 & -3.81 & 94.46 \\
        \textsc{Ellipsis} & 32.81 & 41.78 & 45.33 & 47.89 & 49.37 & \textbf{54.44} & 54.11 & \underline{37.70} & 92.14 \\
        \textsc{FCI Licensing} & 98.02 & 96.33 & 98.16 & 97.98 & \textbf{98.29} & 88.87 & 90.93 & 0.28 & 98.57 \\
        \textsc{Nominal Exp.} & 91.77 & 96.21 & 96.40 & \textbf{97.06} & 96.39 & 92.27 & 93.76 & -4.79 & 92.27 \\
        \textsc{NPI Licensing} & 59.65 & 68.53 & 76.52 & 81.32 & \textbf{84.58} & 73.78 & 74.89 & 8.75 & 93.33 \\
        \textsc{Passive} & 62.20 & 66.11 & 72.04 & 77.47 & 78.36 & 78.17 & \textbf{79.81} & \underline{15.19} & 95.00 \\
        \textsc{Quantifiers} & 54.11 & 53.94 & 52.67 & 50.28 & 53.89 & 56.50 & \textbf{68.33} & \underline{28.10} & 96.43 \\
        \textsc{Question} & 85.08 & 94.81 & 96.80 & 97.35 & \textbf{97.62} & 94.76 & 93.76 & -0.14 & 97.48 \\
        \textsc{Relativization} & 96.81 & 99.50 & 98.50 & 99.33 & \textbf{99.53} & 97.92 & 97.50 & -9.35 & 90.18 \\
        \textsc{Topicalization} & 96.69 & 98.36 & 98.39 & \textbf{99.17} & 99.06 & 96.67 & 94.33 & -1.67 & 97.50 \\
        \textsc{Verb Phrase} & 91.13 & 95.06 & 95.45 & \textbf{95.52} & 95.10 & 93.48 & 93.29 & -1.64 & 93.88 \\\midrule
        \textsc{Overall$_{\text{SLLN-LP}}$} & 79.12 & 83.94 & 87.01 & 89.10 & \textbf{89.99} & 86.68 & 87.18 & 4.62 & 94.61 \\
        \textsc{Overall$_{\text{LP}}$} & 78.44 & 83.14 & 85.57 & 87.16 & \textbf{88.40} & 86.38 & 86.87 & 6.21 & 94.61 \\
        \textsc{Overall$_{\text{MLP}}$} & 72.77 & 79.00 & 82.78 & 84.83 & \textbf{85.38} & 82.68 & 83.05 & 9.23 & 94.61 \\
    \bottomrule
    \end{tabular}}
    \caption{Accuracy of Zh-pythia and Qwen2.5 models on ZhoBLiMP, broken down into different linguistic phenomena. Rows at the bottom compare overall results for different linking functions.}
    \label{tab:assess:overall}
\end{table*}

\begin{figure*}[ht]
    \centering
    \includegraphics[width=\textwidth]{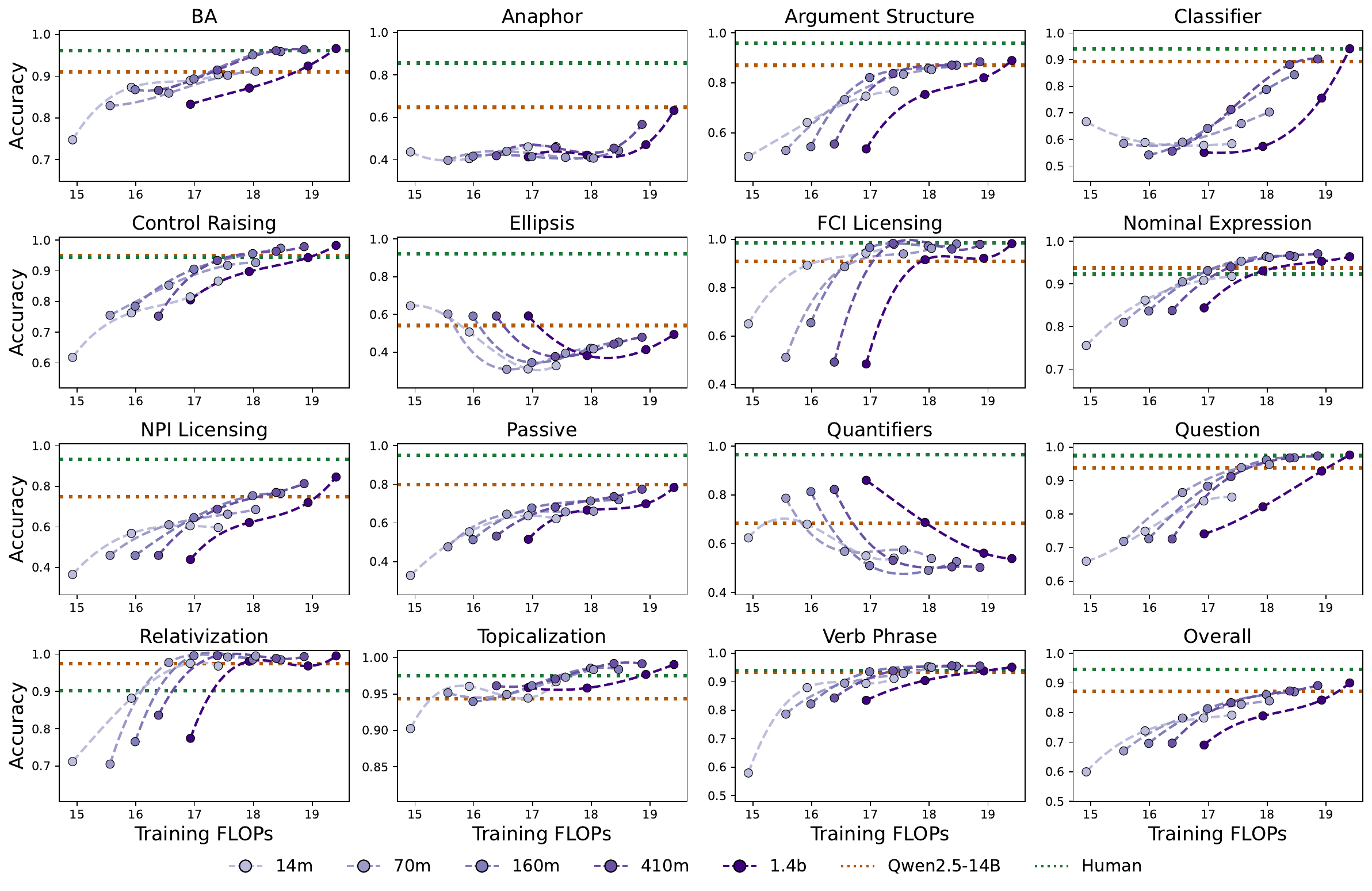}
    \caption{Phenomenon-specific accuracy on ZhoBLiMP plotted against training FLOPs (log scale). Each point represents a distinct Zh-Pythia LM, with models of identical parameter sizes shown in the same color. Points connected by dotted lines represent models of the same size, where higher training FLOPs indicate larger volumes of training tokens. We also plot the performance of Qwen2.5-14B and human as references.}
    \label{fig:asssess:scaling}
\end{figure*}

\begin{figure*}[ht]
    \centering
    \includegraphics[width=\textwidth]{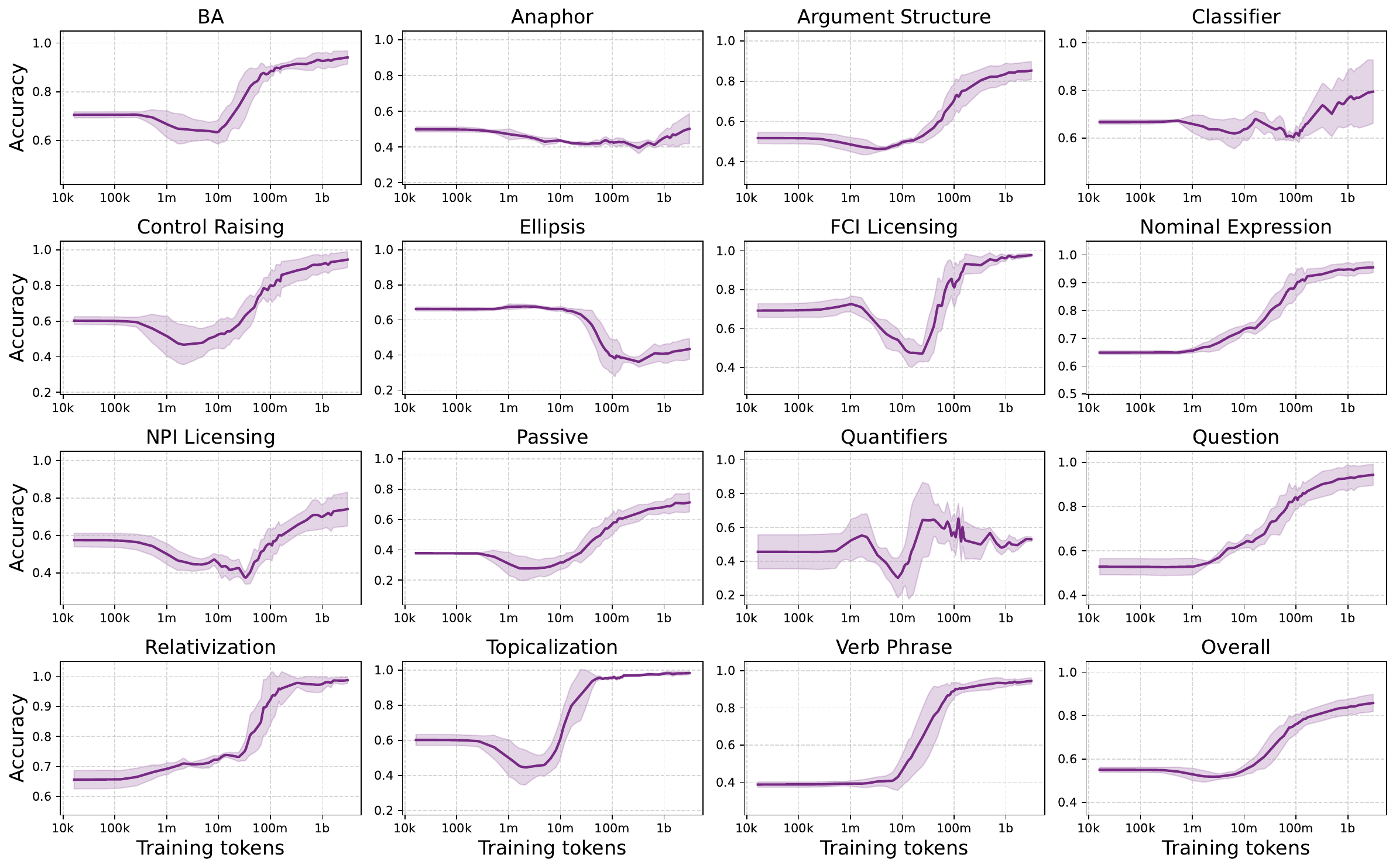}
    \caption{Learning curves across training tokens (calculated from training steps). We analyze 47 intermediate checkpoints from Zh-Pythia models trained on 3B tokens. The plot shows average accuracy across 15 LMs (5 model sizes × 3 seeds) with interpolation smoothing. The shaded area represents the standard deviation among checkpoints at equivalent training volumes.}
    \label{fig:asssess:curves}
\end{figure*}

The scale of LMs can influence the optimal degree of normalization. ~\citet{tjuatja-etal-2025-goes-morcela} demonstrate that the optimal degree of frequency normalization varies across scales. We investigate whether the length normalization varies across scales as well and examine SLLN-LP's applicability to languages other than Chinese. We analyze model predictions using LP, SLLN-LP, and mean LP (with $\alpha \in \left[0, 1\right]$), examining how \textit{accuracy} and $\Delta_{acc}$ respond to different $\alpha$ values.

\begin{itemize}[leftmargin=*]
    \setlength{\itemsep}{0pt}
    \item \textit{Scale analysis}: We evaluate 20 Zh-Pythia LMs, using \textit{cllama} tokenizer, on ZhoBLiMP.
    \item \textit{Cross-linguistic analysis}: We evaluate GPT2-small-Dutch~\citep{de-vries-nissim-2021-good} on BLiMP-NL~\citep{blimp-nl}, Japanese-GPT2-small~\citep{sawada-etal-2024-release} on JBLiMP~\citep{someya2023jblimp}, and Pythia-160m~\citep{biderman2023pythia} on BLiMP~\citep{warstadt-et-al-2020-blimp}.
\end{itemize}

The left of Figure~\ref{fig:debias:scale-language} shows consistent trends in both \textit{accuracy} and $\Delta_{acc}$ across scales, with optimal $\alpha$ between 0.4 and 0.6, yielding the highest \textit{accuracy} and the lowest $\Delta_{acc}$ on ZhoBLiMP.

Unequal lengths in minimal pairs are also common in benchmarks other than ZhoBLiMP: 55.59\% of JBLiMP (Japanese) pairs, 14.37\% of BLiMP (English) pairs, and 20.01\% of BLiMP-NL (Dutch) pairs exhibit unequal lengths.
The length normalization parameter $\alpha$ influences both \textit{accuracy} and $\Delta_{acc}$, though trends vary across languages (see the right plot of Figure~\ref{fig:debias:scale-language}). Japanese shows similarity to Chinese, with \textit{accuracy} first increasing then decreasing, while $\Delta_{acc}$ follows an inverse pattern (yet the trend is not stable due to the comparatively small sample size of JBLiMP). For English, the optimal $\alpha$ falls between 0.2 and 0.4, while for  Dutch it is $\alpha = 1.0$. The results suggest that SLLN-LP is applicable to JBLiMP and English as well, while mean LP provides better length normalization for BLiMP-NL. 

Through comprehensive validation of SLLN-LP in Chinese across tokenizers and model scales, we observe that setting $\alpha = 0.5$ consistently yields robust performance and debiased length normalization. We hypothesize that this optimal degree of length normalization is primarily determined by data- or language-specific characteristics rather than model architecture or size, which seems to be different from the frequency normalization~\citep{tjuatja-etal-2025-goes-morcela}.

Given the above observations, we will use SLLN-LP ($\alpha=0.5$) to assess LMs on ZhoBLiMP for the following reasons:
    (1) SLLN-LP effectively mitigates the evaluation bias arising from length differences,
    demonstrating greater generalizability across tokenizers compared to SLOR and MORCELA.
    (2) SLLN-LP also maintains a stable optimal $\alpha$ value (0.5) across model scales. This consistency allows a fair comparison among the Zh-Pythia checkpoints using a uniform $\alpha$.
\section{Assessing LMs with ZhoBLiMP}
\label{sec:assess}

With \zb{}, Zh-Pythia LMs, and debiased linking functions, we are ready to investigate how LMs acquire Chinese syntax.
We present the assessment in three parts: (1) performance on ZhoBLiMP of five Zh-Pythia LMs trained on 3B tokens and two off-the-shelf LLMs, Qwen2.5-7B and Qwen2.5-14B~\citep{qwen2.5}, (2) analysis on performance variation against training FLOPs, and (3) learning curves of Chinese grammar by examining hundreds of intermediate checkpoints.

\subsection{Overall performance}


The results of the five Zh-Pythia LMs and two pre-trained Qwen2.5 LLMs are presented in Table~\ref{tab:assess:overall}. We first observe that with SLLN-LP, all models achieve higher accuracy, and therefore, the following analysis is based on SLLN-LP.

Model performance ranges from 79\% to almost 90\%. Even the smallest model, Zh-Pythia-14M, achieves 79.12\% accuracy, approximately 30 percentage points above the chance level. Within the Zh-Pythia suite, performance consistently improves with increasing model size. Notably, Zh-Pythia-1.4B achieves the highest performance, outperforming the substantially larger Qwen2.5-7B and Qwen2.5-14B by 2-3 percentage points.

While most of the linguistic knowledge (covered in ZhoBLiMP) can be easily acquired, there are still linguistic phenomena that are challenging even for state-of-the-art LLMs like Qwen2.5. For \textsc{Anaphor}, \textsc{Ellipsis}, \textsc{Passive} and \textsc{Quantifiers}, differences greater than 15 points between the best model and humans are observed. Interestingly, for these four phenomena, Qwen2.5 outperforms Zh-Pythia, suggesting that for more difficult phenomena, a larger training data size and model size are helpful.

\subsection{Effect of scaling}

To analyze how scaling affects model performance across different linguistic phenomena in ZhoBLiMP, we plot performance against training FLOPs for the Zh-Pythia models (see Figure~\ref{fig:asssess:scaling}). Training FLOPs are calculated using the formula $C=6ND$~\citep{hoffman-etal-2022-training}, where $C$ represents training FLOPs, $N$ denotes parameter count, and $D$ indicates training token volume.

For models with 14-160M parameters, overall performance improvements plateau with increased training FLOPs. However, models with 410M and 1.4B parameters continue to show performance gains with additional training tokens, in many phenomena such as \textsc{BA}, \textsc{Argument Structure}, \textsc{NPI}, suggesting that larger Zh-Pythia models may not have reached their full potential and could benefit from extended training. 

Learning curve patterns vary across phenomena. Some phenomena, including \textsc{FCI Licensing}, \textsc{Relativization}, \textsc{Topicalization}, and \textsc{Verb Phrase}, are acquired rapidly within $10^{16}$-$10^{17}$ FLOPs. Others, such as \textsc{BA}, \textsc{Classifier}, \textsc{Nominal expression}, and \textsc{Question}, show more gradual improvement, with performance plateauing around $10^{18}$ FLOPs.

The three challenging phenomena—\textsc{Anaphor}, \textsc{Ellipsis}, and \textsc{Quantifiers}—exhibit different patterns. \textsc{Anaphor} remains near chance level until reaching $10^{18}$ FLOPs. \textsc{Ellipsis} shows initial performance degradation followed by improvement with increased scale. \textsc{Quantifiers}, however, demonstrates consistent performance deterioration as model scale increases. 

\subsection{Learning curves}

To investigate whether LMs acquire syntax gradually or abruptly, we analyze 705 intermediate checkpoints (47 checkpoints $\times$ 5 parameter sizes $\times$ 3 seeds) from our Zh-Pythia models. Figure~\ref{fig:asssess:curves} displays the learning curves for each phenomenon. Most phenomena show minimal improvement during the first 100M training tokens, followed by a sharp performance increase. Performance typically saturates around 1B tokens, consistent across model scales and showing low variance.

Notably, U-shaped learning curves are observed across multiple phenomena, such as \textsc{FCI Licensing} and \textsc{Topicalization}. These curves show initial performance decline around 100M tokens before subsequent improvement. While U-shaped learning has been documented in learning the past tense in English, where LMs might over-generalize the rules of inflection to irregular verbs~\citep{pasttense1987,plunkett1991u,haga-etal-2024-modeling}. Future research can investigate what causes the U-shaped learning in Chinese.

\section{Discussion}
\label{sec:discussion}

This work makes two primary contributions: (1) advancing our understanding of Chinese syntax acquisition through the development of the ZhoBLiMP benchmark and Zh-Pythia LMs, and (2) identifying unequal length bias in minimal pairs and introducing SLLN-LP as a mitigation strategy.
We now summarize our findings 
and discuss their implications.

\subsection{Acquisition of Chinese syntax}

Our assessment reveals that while Chinese syntax can be acquired by small LMs with limited training data, three phenomena remain particularly challenging: \textsc{Anaphor}, \textsc{Ellipsis}, and \textsc{Quantifiers}. 
Performance on these phenomena lags approximately 20 percentage points behind human and exhibits distinct learning patterns across model scales and training steps.
Unlike other phenomena, these three do not show consistent improvement with increased parameter size or training data in the Zh-Pythia suite (see Figure~\ref{fig:asssess:scaling} and~\ref{fig:asssess:curves}). However, the larger Qwen2.5 models achieve better (though still modest) accuracy (see Table~\ref{tab:assess:overall}), suggesting that these phenomena may require greater model capacity and training volume.

Comparing these results with those in English, we find that \textsc{Quantifiers} are challenging in both languages, possibly due to the need for pragmatic and discourse information beyond syntax for quantifier understanding~\citep{Cohen2014SuperlativeQA,sperlich2019syntactic,cremers2022ignorance}.

On the other hand, \textsc{Anaphor} is especially challenging in Chinese but easy in English: Qwen2.5-14B achieves only 64.72\% accuracy on Anaphor in ZhoBLiMP, whereas even GPT-2 reaches 99\% on English BLiMP~\citep{warstadt-et-al-2020-blimp}. 
The lower performance on Chinese anaphora is likely due to the linguistic property of the reflexive pronouns in Chinese: unlike English, Chinese has both 他自己/她自己 (ta-ziji, \textit{himself/herself}) and 自己 (ziji, \textit{self}), with the latter being more frequent. However, in making the minimal pairs, it is only feasible to use the former, as they are gender marked and can be used to control the acceptability of sentences. The status and nature of Chinese anaphora is still debated in theoretical linguistics~\citep{xue1994new-ziji,yu2000chinese-ziji,ZHU2025101267-ziji-taziji}. We believe a more detailed and comprehensive comparison between the two types of reflexive pronouns is needed, along the lines of some recent work assessing LMs' interpretation of Chinese reflexive ziji (\textit{self})~\citep{yang-2025-lm-long-distance-ziji}.

With these findings, we conjecture that the low performance and distinctive learning patterns in these three phenomena stem from their dependence on pragmatic and discourse information, as documented in the Chinese linguistics literature~\citep{huang-1994-anaphora,yuan1998binding,wu2018quantifier,chen2024pragmatic}, distinguishing them from purely syntactic phenomena. 

\subsection{Training small LMs to study syntax acquisition}
Our Zh-Pythia-160M model, trained from scratch on just 1-3B tokens, achieves 87.01\% accuracy, comparable to the much larger Qwen2.5-7B and Qwen2.5-14B. This demonstrates that relatively small LMs can effectively acquire syntactic knowledge, with further scaling yielding limited improvements—only about 3 points of difference between Zh-Pythia-160M and Zh-Pythia-1.4B (see Table~\ref{tab:assess:overall}).

Beyond similar end performance, smaller models exhibit learning patterns comparable to their larger counterparts across various training volumes (see Figure~\ref{fig:asssess:scaling}) and intermediate checkpoints (see Figure~\ref{fig:asssess:curves}). This suggests that studying syntax acquisition patterns in smaller LMs could yield insights similar to those from much larger models.

Our findings align with previous research on English that the effect of scaling is limited for syntax acquisition~\citep{hu-etal-2020-systematic-assessment-syn-gen,zhang-etal-2021-need,warstadt-etal-2023-findings-babylm}. While these studies used models of approximately 100M parameters trained on 100M English words, our Chinese syntax acquisition required an estimated 1-3B Chinese tokens. A key distinction lies in training duration: our models use single-epoch training, whereas English studies typically employ 20 or more epochs. This multiple-epoch approach's effectiveness is corroborated by \citet{wilcox2024bigger}, who demonstrate that repeated exposure to the same 100M words improves BLiMP performance.

Despite these differences, both approaches remain within academic research budgets. We therefore encourage future research on other languages to employ small LMs ($\sim$100M parameters) with small training data (100M-1B tokens) when budgets are limited, as these models achieve adequate performance on \textsc{mpp} benchmarks while exhibiting informative learning patterns for analysis.

\subsection{Minimal pairs, LMs, and linking functions}
We argue that when performing targeted syntactic evaluation, one should consider minimal pairs, LMs, and linking functions wholistically. In previous work~\citep{warstadt-et-al-2020-blimp,rublimp,blimp-nl},  emphasis has been placed on the collection and human validation of minimal pairs. 
The validity and reliability of the linking functions used in evaluation are rarely discussed, as the same function is used uniformly across all models. 

Our findings demonstrate that inappropriate linking functions can introduce evaluation bias. For example, using LP as the linking function leads to unbalanced performance in Zh-Pythia-160m across ZhoBLiMP splits ($\mathcal{D}{=}$: 88.2; $\mathcal{D}{+}$: 60.7; $\mathcal{D}_{-}$: 95.6) (see Table~\ref{tab:debias:fc}). This length-related bias extends beyond ZhoBLiMP to benchmarks in other languages, including BLiMP, JBLiMP and BLiMP-NL.
Functions with more sophisticated normalization, such as SLLN-LP and MORCELA, can mitigate this bias on ZhoBLiMP, reducing $\Delta_{acc}$ to 3.89 and 5.23 while improving accuracy to 86.22 and 85.31. 
These functions show better handling of unequal-length pairs. However, function effectiveness varies across different LMs—SLOR and MORCELA perform well with \textit{cllama} and \textit{word} tokenizers but poorly with \textit{char} tokenizer (see Table~\ref{tab:debias:fc}).

When developing \textsc{mpp} benchmarks, we therefore recommend validating both data quality and the effectiveness of linking functions. This validation should address two key aspects: (1) the compatibility between linking functions and minimal pairs, and (2) the compatibility between linking functions and the target language models.

\section{Conclusion}
\label{sec:conclusion}

In this paper, we introduced \zb{}, the most comprehensive Chinese \textsc{mpp} benchmark, featuring 35k minimal pairs across 15 linguistic phenomena. We trained the Zh-Pythia LM suite---22 models and 230 intermediate checkpoints varying in tokenizers, parameters, and training volumes---and proposed SLLN-LP, a novel linking function for length normalization in \textsc{mpp}. SLLN-LP improves accuracy and reduces ``unequal length'' bias compared to LP and MLP, while matching the performance of SLOR and MORCELA without requiring unigram frequency information.

Our findings reveal two key insights: (1) While small LMs readily acquire most Chinese syntactic phenomena, they struggle with \textsc{Anaphor}, \textsc{Ellipsis}, and \textsc{Quantifiers}, probably due to their unique linguistic properties in Chinese or a reliance on discourse and pragmatic information; (2) Future benchmark development should validate linking functions alongside data quality to ensure fair model evaluation.

\section*{Limitations}
One limitation is that we did not ask human annotators to perform a Likert-scale rating of the sentences, in addition to the force-choice human evaluation we did. Future work can add a Likert-scale rating experiment and compare results of the two task formats. 

Despite our great effort to ensure the quality of the paradigms in ZhoBLiMP, a very small portion of the acceptable sentences may still sound unnatural to some speakers, likely due to the nature of templates-and-vocabulary-based generation in BLiMP-style corpora in general. 
For these rare cases, we ensured that the unacceptable sentences are (much) worse than the acceptable ones. As it is a forced-choice task, this will ensure the soundness of the results.
Future work can explore other methods of minimal pair generation, such as employing an LLM, as demonstrated in \citet{blimp-nl}.

As pointed out by a reviewer, ZhoBLiMP also contains some semantically implausible sentences rather than strictly syntactic ill-formed sentences. Future research can try to tease them apart and model them independently.

\section*{Acknowledgments}
We want to thank the anonymous reviewers and our action editor for their valuable comments and suggestions. 
This work is funded by the Humanities and Social Sciences Grant from the Chinese Ministry of Education (No.~22YJC740020) awarded to Hai Hu, the Shanghai Pujiang Program Grant (No.~22PJC063) awarded to Hai Hu and the General Program of National Natural Science Foundation of China (62176153) awarded to Rui Wang.

\bibliography{tacl2021}

@article{linzen2016assessing,
    title = "Assessing the Ability of {LSTM}s to Learn Syntax-Sensitive Dependencies",
    author = "Linzen, Tal  and
      Dupoux, Emmanuel  and
      Goldberg, Yoav",
    editor = "Lee, Lillian  and
      Johnson, Mark  and
      Toutanova, Kristina",
    journal = "Transactions of the Association for Computational Linguistics",
    volume = "4",
    year = "2016",
    address = "Cambridge, MA",
    publisher = "MIT Press",
    url = "https://aclanthology.org/Q16-1037/",
    doi = "10.1162/tacl_a_00115",
    pages = "521--535"
}

@inproceedings{song2022sling,
    title = "{SLING}: {S}ino Linguistic Evaluation of Large Language Models",
    author = "Song, Yixiao  and
      Krishna, Kalpesh  and
      Bhatt, Rajesh  and
      Iyyer, Mohit",
    editor = "Goldberg, Yoav  and
      Kozareva, Zornitsa  and
      Zhang, Yue",
    booktitle = "Proceedings of the 2022 Conference on Empirical Methods in Natural Language Processing",
    month = dec,
    year = "2022",
    address = "Abu Dhabi, United Arab Emirates",
    publisher = "Association for Computational Linguistics",
    url = "https://aclanthology.org/2022.emnlp-main.305/",
    doi = "10.18653/v1/2022.emnlp-main.305",
    pages = "4606--4634"
}

@inproceedings{someya2023jblimp,
    title = "{JBL}i{MP}: {J}apanese Benchmark of Linguistic Minimal Pairs",
    author = "Someya, Taiga  and
      Oseki, Yohei",
    editor = "Vlachos, Andreas  and
      Augenstein, Isabelle",
    booktitle = "Findings of the Association for Computational Linguistics: EACL 2023",
    month = may,
    year = "2023",
    address = "Dubrovnik, Croatia",
    publisher = "Association for Computational Linguistics",
    url = "https://aclanthology.org/2023.findings-eacl.117/",
    doi = "10.18653/v1/2023.findings-eacl.117",
    pages = "1581--1594"
}

@inproceedings{wilcox2018rnn,
    title = "What do {RNN} Language Models Learn about Filler{--}Gap Dependencies?",
    author = "Wilcox, Ethan  and
      Levy, Roger  and
      Morita, Takashi  and
      Futrell, Richard",
    editor = "Linzen, Tal  and
      Chrupa{\l}a, Grzegorz  and
      Alishahi, Afra",
    booktitle = "Proceedings of the 2018 {EMNLP} Workshop {B}lackbox{NLP}: Analyzing and Interpreting Neural Networks for {NLP}",
    month = nov,
    year = "2018",
    address = "Brussels, Belgium",
    publisher = "Association for Computational Linguistics",
    url = "https://aclanthology.org/W18-5423/",
    doi = "10.18653/v1/W18-5423",
    pages = "211--221"
}

@inproceedings{xiang-et-al-2021-climp,
  author       = {Beilei Xiang and
                  Changbing Yang and
                  Yu Li and
                  Alex Warstadt and
                  Katharina Kann},
  editor       = {Paola Merlo and
                  J{\"{o}}rg Tiedemann and
                  Reut Tsarfaty},
  title        = {{CLiMP}: {A} Benchmark for {Chinese} Language Model Evaluation},
  booktitle    = {Proceedings of the 16th Conference of the European Chapter of the
                  Association for Computational Linguistics: Main Volume, {EACL} 2021,
                  Online, April 19 - 23, 2021},
  pages        = {2784--2790},
  publisher    = {Association for Computational Linguistics},
  year         = {2021},
  url          = {https://doi.org/10.18653/v1/2021.eacl-main.242},
  doi          = {10.18653/v1/2021.eacl-main.242}
}

@article{warstadt-et-al-2020-blimp,
    title = "{BL}i{MP}: The Benchmark of Linguistic Minimal Pairs for {E}nglish",
    author = "Warstadt, Alex  and
      Parrish, Alicia  and
      Liu, Haokun  and
      Mohananey, Anhad  and
      Peng, Wei  and
      Wang, Sheng-Fu  and
      Bowman, Samuel R.",
    editor = "Johnson, Mark  and
      Roark, Brian  and
      Nenkova, Ani",
    journal = "Transactions of the Association for Computational Linguistics",
    volume = "8",
    year = "2020",
    address = "Cambridge, MA",
    publisher = "MIT Press",
    url = "https://aclanthology.org/2020.tacl-1.25/",
    doi = "10.1162/tacl_a_00321",
    pages = "377--392"
}

@article{xue2005penn,
  title={The {Penn} {Chinese} {T}reebank: Phrase structure annotation of a large corpus},
  author={Xue, Naiwen and Xia, Fei and Chiou, Fu-Dong and Palmer, Marta},
  journal={Natural Language Engineering},
  volume={11},
  number={2},
  pages={207--238},
  year={2005},
  publisher={Cambridge University Press},
  doi={https://doi.org/10.1017/S135132490400364X}
}

@article{chinese-llama-alpaca,
      title={Efficient and Effective Text Encoding for {Chinese LLaMA and Alpaca}}, 
      author={Cui, Yiming and Yang, Ziqing and Yao, Xin},
      journal={arXiv preprint arXiv:2304.08177v3},
      url={https://arxiv.org/abs/2304.08177v3},
      year={2023},
      doi={https://doi.org/10.48550/arxiv.2304.08177}
}

@misc{qwen2.5,
    title = {Qwen2.5: A Party of Foundation Models},
    url = {https://qwenlm.github.io/blog/qwen2.5/},
    author = {Team Qwen},
    month = {September},
    year = {2024}
}

@inproceedings{biderman2023pythia,
  title={Pythia: A suite for analyzing large language models across training and scaling},
  author={Biderman, Stella and Schoelkopf, Hailey and Anthony, Quentin Gregory and Bradley, Herbie and O’Brien, Kyle and Hallahan, Eric and Khan, Mohammad Aflah and Purohit, Shivanshu and Prashanth, USVSN Sai and Raff, Edward and others},
  booktitle={International Conference on Machine Learning},
  pages={2397--2430},
  year={2023},
  series={Proceedings of Machine Learning Research},
  url={https://proceedings.mlr.press/v202/biderman23a.html}}

@inproceedings{zhang-etal-2021-need,
    title = "When Do You Need Billions of Words of Pretraining Data?",
    author = "Zhang, Yian  and
      Warstadt, Alex  and
      Li, Xiaocheng  and
      Bowman, Samuel R.",
    editor = "Zong, Chengqing  and
      Xia, Fei  and
      Li, Wenjie  and
      Navigli, Roberto",
    booktitle = "Proceedings of the 59th Annual Meeting of the Association for Computational Linguistics and the 11th International Joint Conference on Natural Language Processing (Volume 1: Long Papers)",
    month = aug,
    year = "2021",
    address = "Online",
    publisher = "Association for Computational Linguistics",
    url = "https://aclanthology.org/2021.acl-long.90",
    doi = "10.18653/v1/2021.acl-long.90",
    pages = "1112--1125",
}

@book{SyntaxHLL,
author="C.-T. James Huang and Y.-H. Audrey Li and Yafei, Li",
title="The Syntax of Chinese",
publisher="Cambridge University Press",
year="2009",
month="03",
DOI="10.1017/cbo9781139166935",
URL="https://cir.nii.ac.jp/crid/1362262946449648000"
}

@inproceedings{haga-etal-2024-modeling,
    title = "Modeling Overregularization in Children with Small Language Models",
    author = "Haga, Akari  and
      Sugawara, Saku  and
      Fukatsu, Akiyo  and
      Oba, Miyu  and
      Ouchi, Hiroki  and
      Watanabe, Taro  and
      Oseki, Yohei",
    editor = "Ku, Lun-Wei  and
      Martins, Andre  and
      Srikumar, Vivek",
    booktitle = "Findings of the Association for Computational Linguistics ACL 2024",
    month = aug,
    year = "2024",
    address = "Bangkok, Thailand and virtual meeting",
    publisher = "Association for Computational Linguistics",
    url = "https://aclanthology.org/2024.findings-acl.865",
    doi = "10.18653/v1/2024.findings-acl.865",
    pages = "14532--14550",
}

@inproceedings{evanson-etal-2023-language,
    title = "Language acquisition: Do children and language models follow similar learning stages?",
    author = "Evanson, Linnea  and
      Lakretz, Yair  and
      King, Jean R{\'e}mi",
    editor = "Rogers, Anna  and
      Boyd-Graber, Jordan  and
      Okazaki, Naoaki",
    booktitle = "Findings of the Association for Computational Linguistics: ACL 2023",
    month = jul,
    year = "2023",
    address = "Toronto, Canada",
    publisher = "Association for Computational Linguistics",
    url = "https://aclanthology.org/2023.findings-acl.773",
    doi = "10.18653/v1/2023.findings-acl.773",
    pages = "12205--12218",
}

@article{plunkett1991u,
  title={U-shaped learning and frequency effects in a multi-layered perception: Implications for child language acquisition},
  author={Plunkett, Kim and Marchman, Virginia},
  journal={Cognition},
  volume={38},
  number={1},
  pages={43--102},
  year={1991},
  doi={https://doi.org/10.1016/0010-0277(91)90022-V},
  publisher={Elsevier}
}

@book{schutze2016empirical,
author = {Schütze, Carson T.},
title = {{The} {E}mpirical {B}ase of {L}inguistics: {Grammaticality} {J}udgments and {L}inguistic {M}ethodology},
year = {2016},
series = {Classics in Linguistics},
number = {2},
address = {Berlin},
publisher = {Language Science Press},
doi = {10.17169/langsci.b89.100}
}

@inproceedings{choshen-etal-2022-grammar,
    title = "The Grammar-Learning Trajectories of Neural Language Models",
    author = "Choshen, Leshem  and
      Hacohen, Guy  and
      Weinshall, Daphna  and
      Abend, Omri",
    editor = "Muresan, Smaranda  and
      Nakov, Preslav  and
      Villavicencio, Aline",
    booktitle = "Proceedings of the 60th Annual Meeting of the Association for Computational Linguistics (Volume 1: Long Papers)",
    month = may,
    year = "2022",
    address = "Dublin, Ireland",
    publisher = "Association for Computational Linguistics",
    url = "https://aclanthology.org/2022.acl-long.568",
    doi = "10.18653/v1/2022.acl-long.568",
    pages = "8281--8297",
}

@article{leong2023bhasa,
      title={{BHASA}: A Holistic {Southeast Asian} Linguistic and Cultural Evaluation Suite for Large Language Models}, 
      author={Wei Qi Leong and Jian Gang Ngui and Yosephine Susanto and Hamsawardhini Rengarajan and Kengatharaiyer Sarveswaran and William Chandra Tjhi},
      year={2023},
      journal={arXiv preprint arXiv:2309.06085v2},
      doi={https://doi.org/10.48550/arxiv.2309.06085}
}

@inproceedings{rublimp,
    title = "{R}u{BL}i{MP}: {R}ussian Benchmark of Linguistic Minimal Pairs",
    author = "Taktasheva, Ekaterina  and
      Bazhukov, Maxim  and
      Koncha, Kirill  and
      Fenogenova, Alena  and
      Artemova, Ekaterina  and
      Mikhailov, Vladislav",
    editor = "Al-Onaizan, Yaser  and
      Bansal, Mohit  and
      Chen, Yun-Nung",
    booktitle = "Proceedings of the 2024 Conference on Empirical Methods in Natural Language Processing",
    month = nov,
    year = "2024",
    address = "Miami, Florida, USA",
    publisher = "Association for Computational Linguistics",
    url = "https://aclanthology.org/2024.emnlp-main.522/",
    doi = "10.18653/v1/2024.emnlp-main.522",
    pages = "9268--9299"
}

@inproceedings{hu-levy-2023-prompting,
    title = "Prompting is not a substitute for probability measurements in large language models",
    author = "Hu, Jennifer  and
      Levy, Roger",
    editor = "Bouamor, Houda  and
      Pino, Juan  and
      Bali, Kalika",
    booktitle = "Proceedings of the 2023 Conference on Empirical Methods in Natural Language Processing",
    month = dec,
    year = "2023",
    address = "Singapore",
    publisher = "Association for Computational Linguistics",
    url = "https://aclanthology.org/2023.emnlp-main.306/",
    doi = "10.18653/v1/2023.emnlp-main.306",
    pages = "5040--5060"
}

@inproceedings{warstadt-etal-2023-findings-babylm,
    title = "Findings of the {B}aby{LM} Challenge: Sample-Efficient Pretraining on Developmentally Plausible Corpora",
    author = "Warstadt, Alex  and
      Mueller, Aaron  and
      Choshen, Leshem  and
      Wilcox, Ethan  and
      Zhuang, Chengxu  and
      Ciro, Juan  and
      Mosquera, Rafael  and
      Paranjabe, Bhargavi  and
      Williams, Adina  and
      Linzen, Tal  and
      Cotterell, Ryan",
    editor = "Warstadt, Alex  and
      Mueller, Aaron  and
      Choshen, Leshem  and
      Wilcox, Ethan  and
      Zhuang, Chengxu  and
      Ciro, Juan  and
      Mosquera, Rafael  and
      Paranjabe, Bhargavi  and
      Williams, Adina  and
      Linzen, Tal  and
      Cotterell, Ryan",
    booktitle = "Proceedings of the BabyLM Challenge at the 27th Conference on Computational Natural Language Learning",
    month = dec,
    year = "2023",
    address = "Singapore",
    publisher = "Association for Computational Linguistics",
    url = "https://aclanthology.org/2023.conll-babylm.1",
    doi = "10.18653/v1/2023.conll-babylm.1",
    pages = "1--34",
}

@inproceedings{hu-etal-2020-systematic-assessment-syn-gen,
    title = "A Systematic Assessment of Syntactic Generalization in Neural Language Models",
    author = "Hu, Jennifer  and
      Gauthier, Jon  and
      Qian, Peng  and
      Wilcox, Ethan  and
      Levy, Roger",
    editor = "Jurafsky, Dan  and
      Chai, Joyce  and
      Schluter, Natalie  and
      Tetreault, Joel",
    booktitle = "Proceedings of the 58th Annual Meeting of the Association for Computational Linguistics",
    month = jul,
    year = "2020",
    address = "Online",
    publisher = "Association for Computational Linguistics",
    url = "https://aclanthology.org/2020.acl-main.158",
    doi = "10.18653/v1/2020.acl-main.158",
    pages = "1725--1744",
}

@article{wilcox2024bigger,
title = {Bigger is not always better: The importance of human-scale language modeling for psycholinguistics},
journal = {Journal of Memory and Language},
volume = {144},
pages = {104650},
year = {2025},
issn = {0749-596X},
doi = {https://doi.org/10.1016/j.jml.2025.104650},
url = {https://www.sciencedirect.com/science/article/pii/S0749596X25000439},
author = {Ethan Gotlieb Wilcox and Michael Y. Hu and Aaron Mueller and Alex Warstadt and Leshem Choshen and Chengxu Zhuang and Adina Williams and Ryan Cotterell and Tal Linzen}
}

@article{lau-et-al-2017-grammar,
author = {Lau, Jey Han and Clark, Alexander and Lappin, Shalom},
title = {Grammaticality, Acceptability, and Probability: A Probabilistic View of Linguistic Knowledge},
journal = {Cognitive Science},
volume = {41},
number = {5},
pages = {1202-1241},
doi = {https://doi.org/10.1111/cogs.12414},
url = {https://onlinelibrary.wiley.com/doi/abs/10.1111/cogs.12414},
eprint = {https://onlinelibrary.wiley.com/doi/pdf/10.1111/cogs.12414},
year = {2017}
}

@article{cremers2022ignorance,
  title={Ignorance implicatures of modified numerals},
  author={Cremers, Alexandre and Coppock, Liz and Dotla{\v{c}}il, Jakub and Roelofsen, Floris},
  journal={Linguistics and Philosophy},
  volume={45},
  pages={683--740},
  year={2022},
  publisher={Springer},
  doi={https://doi.org/10.1007/s10988-021-09336-9}
}

@article{Cohen2014SuperlativeQA,
  title={Superlative quantifiers and meta-speech acts},
  author={Ariel Cohen and Manfred Krifka},
  journal={Linguistics and Philosophy},
  year={2014},
  volume={37},
  pages={41-90},
  pulisher={Springer},
  doi={https://doi.org/10.1007/s10988-014-9144-x}
}

@book{chomsky1965aspects,
  title     = {Aspects of the Theory of Syntax},
  author    = {Chomsky, Noam},
  year      = {1965},
  publisher = {The MIT Press},
  address   = {Cambridge, MA}
}

@article{sperlich2019syntactic,
  title={Syntactic and pragmatic theories of {Chinese} reflexives},
  author={Sperlich, Darcy},
  journal={Lingua},
  volume={221},
  pages={22--36},
  year={2019},
  publisher={Elsevier},
  doi={https://doi.org/10.1016/j.lingua.2019.02.002}
}

@article{chen2024pragmatic,
  title={Pragmatic ellipsis and its pragmatic consequences: Chinese \emph{ye shi} at the syntax-pragmatics interface},
  author={Chen, Yifan and Hu, Xuhui},
  journal={East Asian Pragmatics},
  year={2025},
  volume={9},
  number={3},
  pages={441--462},
  doi={https://doi.org/10.3138/eap.24006}
}

@article{chen2020assessing,
  title={Assessing introspective linguistic judgments quantitatively: The case of \emph{The Syntax of {Chinese}}},
  author={Chen, Zhong and Xu, Yuhang and Xie, Zhiguo},
  journal={Journal of East Asian Linguistics},
  volume={29},
  number={3},
  pages={311--336},
  year={2020},
  publisher={Springer},
  doi={https://doi.org/10.1007/s10831-020-09210-y}
}

@article{blimp-nl,
    author = {Suijkerbuijk, Michelle and Prins, Zoë and Kloots, Marianne de Heer and Zuidema, Willem and Frank, Stefan L.},
    title = {{BLiMP-NL}: A Corpus of {Dutch} Minimal Pairs and Acceptability Judgments for Language Model Evaluation},
    journal = {Computational Linguistics},
    pages = {1-35},
    year = {2025},
    month = {05},
    issn = {0891-2017},
    doi = {10.1162/coli_a_00559},
    url = {https://doi.org/10.1162/coli\_a\_00559},
    eprint = {https://direct.mit.edu/coli/article-pdf/doi/10.1162/coli\_a\_00559/2512113/coli\_a\_00559.pdf},
}

@incollection{pasttense1987,
  title={On learning the past tenses of {English} verbs},
  author={Rumelhart, David E. and McClelland, James L.},
  booktitle={Parallel Distributed Processing},
  volume={2},
  year={1986},
  publisher={MIT Press},
}

@article{gage1994new,
  title={A new algorithm for data compression},
  author={Gage, Philip},
  journal={The C Users Journal},
  volume={12},
  number={2},
  pages={23--38},
  year={1994},
  publisher={R \& D Publications, Inc. Lawrence, KS, USA}
}

@inproceedings{sennrich-etal-2016-neural,
    title = "Neural Machine Translation of Rare Words with Subword Units",
    author = "Sennrich, Rico  and
      Haddow, Barry  and
      Birch, Alexandra",
    editor = "Erk, Katrin  and
      Smith, Noah A.",
    booktitle = "Proceedings of the 54th Annual Meeting of the Association for Computational Linguistics (Volume 1: Long Papers)",
    month = aug,
    year = "2016",
    address = "Berlin, Germany",
    publisher = "Association for Computational Linguistics",
    url = "https://aclanthology.org/P16-1162/",
    doi = "10.18653/v1/P16-1162",
    pages = "1715--1725"
}

@article{llama-2023,
  author       = {Hugo Touvron and
                  Thibaut Lavril and
                  Gautier Izacard and
                  Xavier Martinet and
                  Marie{-}Anne Lachaux and
                  Timoth{\'{e}}e Lacroix and
                  Baptiste Rozi{\`{e}}re and
                  Naman Goyal and
                  Eric Hambro and
                  Faisal Azhar and
                  Aur{\'{e}}lien Rodriguez and
                  Armand Joulin and
                  Edouard Grave and
                  Guillaume Lample},
  title        = {{LLaMA}: Open and Efficient Foundation Language Models},
  journal      = {arXiv preprint arXiv:2302.13971v1},
  year         = {2023},
  doi          = {https://doi.org/10.48550/arxiv.2302.13971},
}

@article{hu2025bilingual,
  title={Bilingual influences and sources of variability in acceptability judgments: A case study of {Chinese}},
  author={Hu, Hai and Li, Aini and Patterson, Yina and Huang, Jiahui and Lin, Chien-Jer Charles},
  journal={Lingua},
  volume={318},
  pages={103911},
  year={2025},
  doi={https://doi.org/10.1016/j.lingua.2025.103911},
  publisher={Elsevier}
}

@inproceedings{tjuatja-etal-2025-goes-morcela,
    title = "What Goes Into a {LM} Acceptability Judgment? {Rethinking} the Impact of Frequency and Length",
    author = "Tjuatja, Lindia  and
      Neubig, Graham  and
      Linzen, Tal  and
      Hao, Sophie",
    editor = "Chiruzzo, Luis  and
      Ritter, Alan  and
      Wang, Lu",
    booktitle = "Proceedings of the 2025 Conference of the Nations of the Americas Chapter of the Association for Computational Linguistics: Human Language Technologies (Volume 1: Long Papers)",
    month = apr,
    year = "2025",
    address = "Albuquerque, New Mexico",
    publisher = "Association for Computational Linguistics",
    url = "https://aclanthology.org/2025.naacl-long.109/",
    doi = "10.18653/v1/2025.naacl-long.109",
    pages = "2173--2186",
    ISBN = "979-8-89176-189-6"
}

@inproceedings{de-vries-nissim-2021-good,
    title = "As Good as New. {How} to Successfully Recycle {E}nglish {GPT}-2 to Make Models for Other Languages",
    author = "de Vries, Wietse  and
      Nissim, Malvina",
    editor = "Zong, Chengqing  and
      Xia, Fei  and
      Li, Wenjie  and
      Navigli, Roberto",
    booktitle = "Findings of the Association for Computational Linguistics: ACL-IJCNLP 2021",
    month = aug,
    year = "2021",
    address = "Online",
    publisher = "Association for Computational Linguistics",
    url = "https://aclanthology.org/2021.findings-acl.74/",
    doi = "10.18653/v1/2021.findings-acl.74",
    pages = "836--846"
}

@inproceedings{sawada-etal-2024-release,
    title = "Release of Pre-Trained Models for the {J}apanese Language",
    author = "Sawada, Kei  and
      Zhao, Tianyu  and
      Shing, Makoto  and
      Mitsui, Kentaro  and
      Kaga, Akio  and
      Hono, Yukiya  and
      Wakatsuki, Toshiaki  and
      Mitsuda, Koh",
    editor = "Calzolari, Nicoletta  and
      Kan, Min-Yen  and
      Hoste, Veronique  and
      Lenci, Alessandro  and
      Sakti, Sakriani  and
      Xue, Nianwen",
    booktitle = "Proceedings of the 2024 Joint International Conference on Computational Linguistics, Language Resources and Evaluation (LREC-COLING 2024)",
    month = may,
    year = "2024",
    address = "Torino, Italia",
    publisher = "ELRA and ICCL",
    url = "https://aclanthology.org/2024.lrec-main.1213/",
    pages = "13898--13905"
}

@article{sprouse2013comparison-linguistic-inquiry,
  title={A comparison of informal and formal acceptability judgments using a random sample from {Linguistic Inquiry} 2001--2010},
  author={Sprouse, Jon and Sch{\"u}tze, Carson T. and Almeida, Diogo},
  journal={Lingua},
  volume={134},
  pages={219--248},
  year={2013},
  publisher={Elsevier},
  doi={https://doi.org/10.1016/j.lingua.2013.07.002}
}

@book{huang-1994-anaphora, place={Cambridge}, series={Cambridge Studies in Linguistics}, title={The Syntax and Pragmatics of Anaphora: A Study with Special Reference to Chinese}, publisher={Cambridge University Press}, author={Huang, Yan}, year={1994}, collection={Cambridge Studies in Linguistics}, doi={https://doi.org/10.1017/CBO9780511554292}}

@article{yuan1998binding,
author = {Boping Yuan},
title ={Interpretation of binding and orientation of the {Chinese} reflexive ziji by
                {English} and {Japanese} speakers},

journal = {Second Language Research},
volume = {14},
number = {4},
pages = {324-340},
year = {1998},
doi = {10.1191/026765898670904111},

URL = { 
    
        https://doi.org/10.1191/026765898670904111
    
    

},
eprint = { 
    
        https://doi.org/10.1191/026765898670904111
    
    

}
}

@article{wu2018quantifier,
title = {Expressing (inter)subjectivity with universal quantification: A pragmatic account of {Plural NP} + dou expressions in {Mandarin Chinese}},
journal = {Journal of Pragmatics},
volume = {128},
pages = {1-21},
year = {2018},
issn = {0378-2166},
doi = {https://doi.org/10.1016/j.pragma.2018.02.003},
url = {https://www.sciencedirect.com/science/article/pii/S0378216617305362},
author = {Haiping Wu and Hongyin Tao}
}

@inproceedings{alkhamissi2025language-cognition-llms-out-grow,
    title = "From Language to Cognition: How {LLM}s Outgrow the Human Language Network",
    author = "AlKhamissi, Badr  and
      Tuckute, Greta  and
      Tang, Yingtian  and
      Binhuraib, Taha Osama A  and
      Bosselut, Antoine  and
      Schrimpf, Martin",
    editor = "Christodoulopoulos, Christos  and
      Chakraborty, Tanmoy  and
      Rose, Carolyn  and
      Peng, Violet",
    booktitle = "Proceedings of the 2025 Conference on Empirical Methods in Natural Language Processing",
    month = nov,
    year = "2025",
    address = "Suzhou, China",
    publisher = "Association for Computational Linguistics",
    url = "https://aclanthology.org/2025.emnlp-main.1237/",
    doi = "10.18653/v1/2025.emnlp-main.1237",
    pages = "24332--24350",
    ISBN = "979-8-89176-332-6"
}

@inproceedings{yang-2025-lm-long-distance-ziji,
    title = "Language Models at the Syntax-Semantics Interface: A Case Study of the Long-Distance Binding of {C}hinese Reflexive Ziji",
    author = "Yang, Xiulin",
    editor = "Rambow, Owen  and
      Wanner, Leo  and
      Apidianaki, Marianna  and
      Al-Khalifa, Hend  and
      Eugenio, Barbara Di  and
      Schockaert, Steven",
    booktitle = "Proceedings of the 31st International Conference on Computational Linguistics",
    month = jan,
    year = "2025",
    address = "Abu Dhabi, UAE",
    publisher = "Association for Computational Linguistics",
    url = "https://aclanthology.org/2025.coling-main.257/",
    pages = "3808--3824"
}

@book{yu2000chinese-ziji,
  title={Chinese {R}eflexives},
  author={Yu, William Xian-fu},
  year={2000},
  publisher={Peeters Publishers},
}

@inproceedings{xue1994new-ziji,
  title={A new perspective on {Chinese} ziji},
  author={Xue, Ping and Pollard, Carl and Sag, Ivan A.},
  booktitle={The Proceedings of the Thirteenth West Coast Conference on Formal Linguistics},
  year={1994}
}

@article{ZHU2025101267-ziji-taziji,
title = {Who is ziji or ta-ziji? {An ERP} study on the processing mechanism of {Chinese} bare and compound reflexives},
journal = {Journal of Neurolinguistics},
volume = {75},
pages = {101267},
year = {2025},
issn = {0911-6044},
doi = {https://doi.org/10.1016/j.jneuroling.2025.101267},
url = {https://www.sciencedirect.com/science/article/pii/S0911604425000235},
author = {Ruoxuan Zhu and Xingsan Chai}
}

@article{hoffman-etal-2022-training,
      title={Training Compute-Optimal Large Language Models}, 
      author={Jordan Hoffmann and Sebastian Borgeaud and Arthur Mensch and Elena Buchatskaya and Trevor Cai and Eliza Rutherford and Diego de Las Casas and Lisa Anne Hendricks and Johannes Welbl and Aidan Clark and Tom Hennigan and Eric Noland and Katie Millican and George van den Driessche and Bogdan Damoc and Aurelia Guy and Simon Osindero and Karen Simonyan and Erich Elsen and Jack W. Rae and Oriol Vinyals and Laurent Sifre},
      year={2022},
      journal={arXiv preprint arXiv:2203.15556v1},
      doi={https://doi.org/10.48550/arxiv.2203.15556}, 
}

@article{lau2020-penlp,
    author = {Lau, Jey Han and Armendariz, Carlos and Lappin, Shalom and Purver, Matthew and Shu, Chang},
    title = {How Furiously Can Colorless Green Ideas Sleep? {Sentence} Acceptability in Context},
    journal = {Transactions of the Association for Computational Linguistics},
    volume = {8},
    pages = {296-310},
    year = {2020},
    month = {06},
    issn = {2307-387X},
    doi = {10.1162/tacl_a_00315},
    url = {https://doi.org/10.1162/tacl_a_00315},
    eprint = {https://direct.mit.edu/tacl/article-pdf/doi/10.1162/tacl_a_00315/1923610/tacl_a_00315.pdf},
}

@article{wu2016google-NMT,
  title={Google's neural machine translation system: Bridging the gap between human and machine translation},
  author={Yonghui Wu and Mike Schuster and Zhifeng Chen and Quoc V. Le and Mohammad Norouzi and Wolfgang Macherey and Maxim Krikun and Yuan Cao and Qin Gao and Klaus Macherey and Jeff Klingner and Apurva Shah and Melvin Johnson and Xiaobing Liu and {Ł}ukasz Kaiser and Stephan Gouws and Yoshikiyo Kato and Taku Kudo and Hideto Kazawa and Keith Stevens and George Kurian and Nishant Patil and Wei Wang and Cliff Young and Jason Smith and Jason Riesa and Alex Rudnick and Oriol Vinyals and Greg Corrado and Macduff Hughes and Jeffrey Dean},
  journal={arXiv preprint arXiv:1609.08144v2},
  year={2016},
  doi={https://doi.org/10.48550/arxiv.1609.08144}
}

@inproceedings{ueda2024token-length-bias,
    title = "Token-length Bias in Minimal-pair Paradigm Datasets",
    author = "Ueda, Naoya  and
      Mita, Masato  and
      Oka, Teruaki  and
      Komachi, Mamoru",
    editor = "Calzolari, Nicoletta  and
      Kan, Min-Yen  and
      Hoste, Veronique  and
      Lenci, Alessandro  and
      Sakti, Sakriani  and
      Xue, Nianwen",
    booktitle = "Proceedings of the 2024 Joint International Conference on Computational Linguistics, Language Resources and Evaluation (LREC-COLING 2024)",
    month = may,
    year = "2024",
    address = "Torino, Italia",
    publisher = "ELRA and ICCL",
    url = "https://aclanthology.org/2024.lrec-main.1410/",
    pages = "16224--16236"
}
\bibliographystyle{acl_natbib}

\appendix
\section{Evaluate linking functions against human Likert-scale ratings}

Having established SLLN-LP's effectiveness for length normalization on ZhoBLiMP and potentially other languages' minimal pairs, we evaluate its ability to predict human gradient acceptability ratings in Chinese. We utilize human ratings from \citet{chen2020assessing} and \citet{hu2025bilingual}, which differ from \zb{} in using expert-crafted minimal pairs from linguistic journals and textbooks. These studies collected Likert scale ratings for individual sentences, with acceptability measured as z-scored averages across multiple annotators.

We analyze correlations in two dimensions: (1) between absolute LM scores $f(x)$ and human ratings of $x$, and (2) between score differences and rating differences across grammatical-ungrammatical sentence pairs. Correlation strength (Pearson's $r$) indicates each function's predictive power  (see Table~\ref{tab:app:ls}).

\begin{table}[ht]
\centering
\resizebox{\linewidth}{!}{
    \begin{tabular}{l|lll|lll}
    \toprule
         & \multicolumn{3}{c|}{Pointwise rating ($r$)} & \multicolumn{3}{c}{Pairwise $\Delta$ ($r$)} \\\midrule
        Func. & cllama & word & char & cllama & word & char \\\midrule
        LP & 32.44 & 25.58 & 24.84 & 44.61 & 35.72 & 36.63 \\
        MLP & 41.79 & 28.58 & 31.12 & 39.51 & 30.77 & 33.96 \\
        SLOR & 46.24 & 32.63 & 32.28 & 48.82 & 38.89 & 38.21 \\
        MCL & 56.35 & 47.76 & 46.76 & 55.19 & 46.43 & 47.24 \\\midrule
        SLLN & 46.69 & 38.68 & 39.05 & 50.94 & 42.97 & 43.97 \\
    \bottomrule
    \end{tabular}}
    \caption{Correlation of human acceptability ratings and different linking functions from Zh-Pythia-160m, trained with different tokenizers. LP: log-probabilities; MLP: mean LP; MCL: MORCELA; SLLN: SLLN-LP ($\alpha=0.5$). Human ratings are from \citet{chen2020assessing} and \citet{hu2025bilingual}.}
    \label{tab:app:ls}
\end{table}

SLLN-LP consistently ranks second in performance across all tokenizers and settings, behind only MORCELA. While MORCELA achieves higher correlations, SLLN-LP provides a simpler conceptualization that doesn't require frequency normalization.

\section{Performance on excluded paradigms}

\begin{table}[ht]
    \centering
    \begin{tabular}{lcc} 
    \toprule
         & included & excluded \\
        \# paradigms & 118 & 11 \\
        Human agreement & 93.9 & 59.6 \\\midrule
        LP & 85.9 & 77.0 \\
        mean LP & 83.7 & 70.7 \\
        SLOR & 86.9 & 79.3 \\
        MORCELA & 88.4 & 80.6 \\
        SLLN-LP & 87.0 & 78.2 \\
    \bottomrule
    \end{tabular}
    \caption{Comparison between model performance (Zh-Pythia-160M) on included paradigms and excluded paradigms.}
    \label{tab:app:excluded}
\end{table}

We exclude 11 paradigms for their low human agreement, which may be caused by poor data quality or ambiguous acceptability. We find that Zh-Pythia-160M has a lower accuracy when evaluated on these excluded paradigms. We will release these paradigms and their human annotations too, to facilitate research on human agreement and LM scores.

\end{CJK*}
\end{document}